\pdfoutput=1

\documentclass[10pt,twocolumn,letterpaper]{article}

\usepackage[pagenumbers]{cvpr}

\usepackage{graphicx}
\usepackage{amsmath}
\usepackage{amssymb}
\usepackage{booktabs}
\usepackage{siunitx}
\usepackage{bm}
\usepackage{subcaption}
\usepackage{soul}

\usepackage[pagebackref,breaklinks,colorlinks]{hyperref}

\usepackage[accsupp]{axessibility}

\usepackage[capitalize]{cleveref}
\crefname{section}{Sec.}{Secs.}
\Crefname{section}{Section}{Sections}
\Crefname{table}{Table}{Tables}
\crefname{table}{Tab.}{Tabs.}

\DeclareMathSymbol{@}{\mathord}{letters}{"3B}
\newcommand{\uarr}{$\uparrow$}
\newcommand{\darr}{$\downarrow$}
\newcommand{\tb}[1]{\textbf{#1}}
\newcommand{\cm}{\checkmark}
\newcommand{\code}[1]{\textbf{\texttt{\small #1}}}

\newcommand{\texthead}[1]{\noindent\textbf{#1}}
\newcommand{\red}[1]{{\color{red} #1}} 

\def\confName{CVPR}
\def\confYear{2022}

\begin{document}

\title{\vspace*{-0.10in}PONI: Potential Functions for ObjectGoal Navigation\\with Interaction-free Learning \vspace*{-0.05in}}

\author{
\textbf{Santhosh Kumar Ramakrishnan$^{1,2}$,  Devendra Singh Chaplot$^{1}$,  Ziad Al-Halah$^{2}$,} \\
\textbf{Jitendra Malik$^{1,3}$, Kristen Grauman$^{1,2}$} \\
$^{1}$Meta AI~~$^{2}$UT Austin~~$^{3}$UC Berkeley
}

\maketitle

\begin{abstract}

State-of-the-art approaches to ObjectGoal navigation (ObjectNav) rely on reinforcement learning and typically require significant computational resources and time for learning. We propose Potential functions for ObjectGoal Navigation with Interaction-free learning (PONI), a modular approach that disentangles the skills of `where to look?' for an object and `how to navigate to $(x, y)$?'. Our key insight is that `where to look?' can be treated purely as a perception problem, and learned without environment interactions. To address this, we propose a network that predicts two complementary potential functions conditioned on a semantic map and uses them to decide where to look for an unseen object.  We train the potential function network using supervised learning on a passive dataset of top-down semantic maps, and integrate it into a modular framework to perform ObjectNav. Experiments on Gibson and Matterport3D demonstrate that our method achieves the state-of-the-art for ObjectNav while incurring up to $1@600\times$ less computational cost for training. Code and pre-trained models are available.\footnote{Website: {\scriptsize\url{https://vision.cs.utexas.edu/projects/poni/}}}

\end{abstract}
\begin{figure*}
    \centering
    \includegraphics[width=\textwidth,trim={0 57cm 33cm 0},clip]{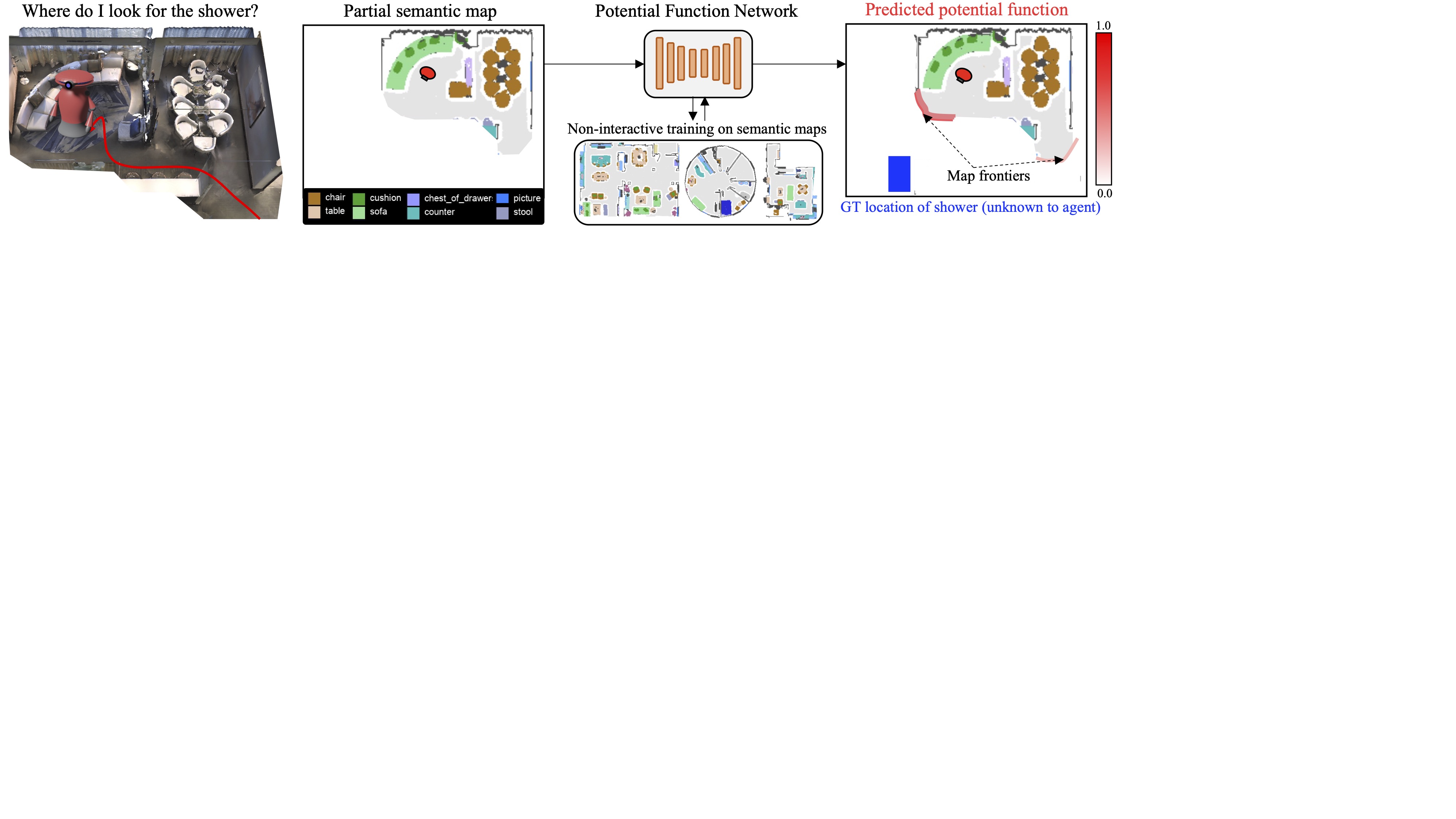}
    \caption{\textbf{Potential functions for ObjectGoal Navigation with Interaction-free learning (PONI)}. We introduce a method to decide `where to look?' for an unseen object in indoor 3D environments. Our key insight is that this is fundamentally a perception problem that can be solved without any interactive learning. We address this problem by defining a \emph{potential function}, which is $[0, 1]$-valued function that represents the value of visiting each location in order to find the object. Given the potential function, we can simply select its argmax location to decide where to look for the object. We propose a potential function network that uses geometric and semantic cues from a partially filled semantic map to predict the potential function for the object of interest (e.g., a shower). We train this network in a non-interactive fashion on a dataset of semantic maps, and integrate it into a modular framework for performing ObjectNav.} 
    \label{fig:intro}
\end{figure*}

\section{Introduction}

Embodied visual navigation is the computer vision problem where an agent uses visual sensing to actively interact with the world and perform navigation tasks~\cite{anderson2018evaluation,savva2019habitat,anderson2018vision,batra2020objectnav}. We have witnessed substantial progress in embodied visual navigation over the past decade, fueled by the availability of large-scale photorealistic 3D scene datasets~\cite{chang2017matterport3d,xia2018gibson,ramakrishnan2021habitat} and fast simulators for embodied navigation~\cite{xia2018gibson,kolve2017ai2,savva2019habitat}. 

ObjectNav has gained popularity in recent years~\cite{savva2017minos,anderson2018evaluation,batra2020objectnav}. Here, an agent enters a novel and unmapped 3D scene, and is given an object category to navigate to (e.g., a chair).  To successfully solve the task, the agent needs to efficiently navigate to the object and stop near it within a given time budget. This is a fundamental problem for embodied agents which requires semantic reasoning about the world (e.g., TV is in the living room, oven is in the kitchen, and chairs are near tables), and it serves as a precursor to more complex object manipulation tasks~\cite{szot2021habitat,kolve2017ai2}.

Prior work has made good progress on this task by formulating it as a reinforcement learning (RL) problem and developing useful representations~\cite{yang2018visual,du2020learning}, auxiliary tasks~\cite{ye2021auxiliary}, data augmentation techniques~\cite{maksymets2021thda}, and improved reward functions~\cite{maksymets2021thda}. Despite this progress, end-to-end RL incurs high computational cost, has poor sample efficiency, and tends to generalize poorly to new scenes~\cite{chaplot2020object,campari2020exploiting,maksymets2021thda} since skills like moving without collisions, exploration, and stopping near the object are all learned from scratch purely using RL. Modular navigation methods aim to address these issues by disentangling `where to look for an object?'  and `how to navigate to $(x, y)$?'~\cite{chaplot2020object,liang2021sscnav}. These methods have emerged as strong competitors to end-to-end RL with good sample efficiency, better generalization to new scenes, and simulation to real-world transfer~\cite{chaplot2019learning,chaplot2020object}. However, since `where to look?' is formulated as an RL problem with interactive reward-based learning, these methods require extensive computational resources ($\sim$8 GPUs) over multiple days for training. 

We hypothesize that the `where to look?' question for an unseen object is fundamentally a perception problem, and can be learned without any interactions. Based on this insight, we introduce a simple-yet-effective method for ObjectNav -- \emph{\textbf{P}otential functions for \textbf{O}bjectGoal \textbf{N}avigation with \textbf{I}nteraction-free learning (PONI).} A potential function is a 0-1 valued function defined at the frontiers of a 2D top-down semantic map\footnote{The 2D semantic map contains the object category per map location.}, i.e., the map locations that lie on the edges of the explored and unexplored regions (see~\cref{fig:intro}, right). It represents the value of visiting each location in order to find the object (higher the value, the better). Given the potential function, we can decide `where to look?' by simply selecting the maximum potential location. 

We propose  the \emph{potential function network}, a convolutional encoder-decoder model that estimates the potential function from a partially filled semantic map. Critically, we propose to train it interaction-free using a dataset of top-down semantic maps  obtained from 3D semantic annotations~\cite{chang2017matterport3d,xia2018gibson} (see~\cref{fig:intro}, center). This is unlike prior work on RL which interactively learns ObjectNav policies by designing reward functions using the same semantic annotations~\cite{chaplot2020object,maksymets2021thda,ye2021auxiliary}. Specifically, our network predicts two complementary potential functions by leveraging geometric and semantic cues in the semantic map (e.g., the environment layout, room-object, and object-object relationships). The \emph{area potential function} captures the unexplored areas in the map for efficient exploration, while the \emph{object potential function} is a geodesic-distance based function that helps decide how to reach the object. Once trained, we deploy the potential function network in a modular framework for ObjectNav, where we combine the area and object potential predictions to decide where to look for a goal object.

We perform experiments on the photorealistic 3D environments of Gibson~\cite{xia2018gibson} and Matterport3D~\cite{chang2017matterport3d}. Our proposed method outperforms a state-of-the-art modular RL method~\cite{chaplot2020object} on Gibson with $7\times$ lower training cost, and a state-of-the-art end-to-end RL method~\cite{maksymets2021thda} on MP3D with $1@600\times$ lower training cost. Our method sets the state-of-the-art on the Habitat ObjectNav challenge leaderboard~\cite{batra2020objectnav} when compared to previously published methods. 
\section{Related Work}

\textbf{Visual navigation.} Prior work has proposed a variety of visual navigation tasks such as PointNav~\cite{anderson2018evaluation,savva2019habitat,savva2017minos}, ObjectNav~\cite{batra2020objectnav,savva2017minos}, RoomNav~\cite{anderson2018evaluation,savva2017minos}, ImageNav~\cite{zhu2017target,savinov2018semi,al-halah2022zsel}, AudioNav~\cite{chen2020soundspaces,chen2021semantic}, instruction following~\cite{anderson2018vision,krantz2020beyond}, and question answering~\cite{das2018embodied,gordon2018iqa}. Research into memory models such as recurrent networks~\cite{hochreiter1997long,wijmans2019dd,savva2019habitat}, metric maps~\cite{chen2018learning,gupta2017cognitive,henriques2018mapnet,ramakrishnan2021exploration}, topological graphs~\cite{savinov2018semi,chaplot2020neural}, and episodic memory~\cite{fang2019scene} has facilitated significant improvements on these tasks. 
In this work, we propose a novel strategy to tackle ObjectNav. \vspace{0.05in}

\textbf{ObjectGoal navigation.} Recent work on end-to-end RL for ObjectNav has proposed improved visual representations~\cite{mousavian2019visual,yang2018visual}, auxiliary tasks~\cite{ye2021auxiliary}, and data augmentation techniques~\cite{maksymets2021thda} to improve generalization to novel scenes. Improved visual representations include semantic segmentations~\cite{mousavian2019visual}, spatial attention maps~\cite{mayo2021visual}, and object relation graphs~\cite{yang2018visual,du2020learning,pal2021learning,moghaddam2021optimistic,druon2020visual,zhang2021hierarchical}. Prior work also learns auxiliary tasks such as predicting agent dynamics, environment states, and map coverage simultaneously with ObjectNav and achieves promising results~\cite{ye2021auxiliary}. Most recently, treasure hunt data augmentation~\cite{maksymets2021thda} achieved state-of-the-art on ObjectNav by training with artificially inserted objects and improving the RL reward~\cite{maksymets2021thda}.

Modular RL methods for ObjectNav have also emerged as strong competitors to end-to-end RL~\cite{chaplot2020object,liang2021sscnav}. They rely on individual modules for semantic mapping, high-level semantic exploration (i.e., where to look?), and low-level navigation (i.e., how to navigate to $(x, y)$?). The semantic exploration module is learned through RL, yet it is more sample-efficient and generalizes better than end-to-end RL due to modularity and shorter time horizons. We propose a novel strategy for the semantic exploration module. Specifically, we decide `where to look?' using a potential function network that is trained non-interactively via supervised learning on a dataset of semantic maps. When integrated into a state-of-the-art modular pipeline~\cite{chaplot2020object}, our method achieves better performance while incurring significantly lower computational cost for training.  

\textbf{Non-interactive learning for navigation.}
Learning from passive (non-interactive) data has emerged as a good recipe for learning navigation policies.  Behavior cloning learns navigation policies from expert action supervision~\cite{gupta2017cognitive,das2018embodied,wijmans2019embodied,anderson2018vision,chen2018learning}, but typically underperforms for complex navigation tasks~\cite{chang2020semantic,chaplot2020neural} and may require subsequent RL finetuning~\cite{chen2018learning,das2018embodied}. Prior work also focuses on pre-training visual representations from image supervision~\cite{sax2020learning,das2018embodied}, environment-level representations from videos~\cite{ramakrishnan2021environment}, and learning navigation subroutines from videos~\cite{kumar2020learning}. However, they require subsequent policy learning to solve specific navigation tasks of interest. Recent work on self-supervised navigation directly learns distance and semantic scoring functions using passive image~\cite{chaplot2020neural} and video~\cite{hahn2021no,chang2020semantic} collections, and uses analytical strategies for ImageNav and ObjectNav. In contrast, we propose a supervised strategy to learn ObjectNav policies from a passive dataset of 2D semantic maps, and demonstrate state-of-the-art performance with high computational efficiency for training. \vspace{0.1in} 

\textbf{Waypoint-based navigation.}
Prior work on waypoint-based navigation repeatedly predicts intermediate goals en route to the target, then uses low-level navigators to reach these intermediate goals
~\cite{chaplot2019learning,chen2020learning,gan2019look,bansal2020combining,nair2019hierarchical}. Such policies can be learned through reinforcement learning~\cite{chen2020learning,chaplot2019learning,ramakrishnan2020occupancy,wu2020spatial,liang2021sscnav}, supervised learning~\cite{stein2018learning,bansal2020combining,gan2019look}, or just analytical planning without any learning~\cite{yamauchi1997frontier}. Potential functions can be interpreted as a Q-value function for ObjectNav~\cite{liang2021sscnav}, but predicted only on frontiers and learned in a supervised fashion from collections of top-down semantic maps. The approach in~\cite{stein2018learning} learns to predict value functions at the frontiers for PointGoal navigation, i.e., to a known $(x, y)$ location, in synthetically generated mazes and lab floor-plans. In contrast, we tackle the more challenging ObjectNav task where the goal location is not known a priori, and we focus on diverse real-world indoor environments where semantic reasoning is required to find the goal (e.g., object-object and object-room relationships). We introduce the area potential function to encourage exploration and information gathering, and the object potential function to perform semantic reasoning. We empirically demonstrate the value of these components for ObjectNav in~\cref{sec:results}. 
\begin{figure*}
    \centering
    \includegraphics[width=\textwidth,trim={0 45.0cm 6cm 0},clip]{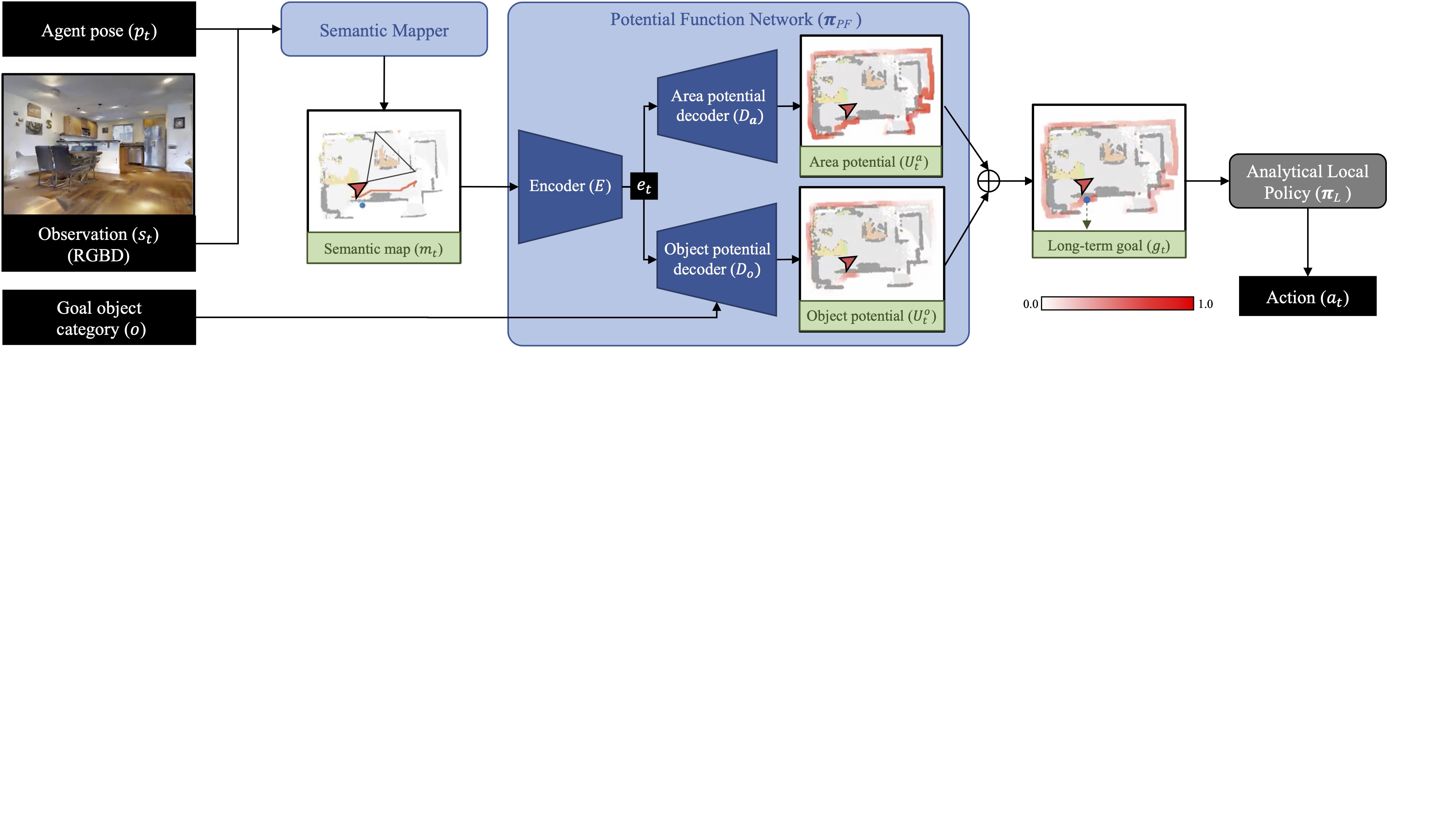}
    \vspace*{-0.25in}
    \caption{\textbf{Architecture for PONI:} Our model consists of three components. The semantic mapper uses RGB-D and pose sensor readings to build an allocentric map $m_t$ of the world. The potential function network $\pi_{pf}$ uses the semantic map and the goal object category $o_t$ to predict the area and object potential functions. The two potentials are averaged and the maximum location is sampled as the long-term goal. The local policy $\pi_L$ navigates the agent towards the long-term goal $g_t$ using analytical path-planning. }
    \label{fig:approach}
\end{figure*}

\section{Approach}

We next formally define the ObjectNav task and introduce our method.

\subsection{ObjectNav Definition}
\label{sec:objectnav_def}

An agent is tasked with navigating to an object specified by its category label (e.g., chair) in an unexplored environment~\cite{batra2020objectnav}. At the start of an episode, the agent is spawned at a random navigable location in the environment. At each time-step $t$, the agent receives $640 \times 480$ RGB-D sensor readings $s_t$, $(x, y, \theta)$ odometer readings, and the goal category $o$. The odometer readings are aggregated over time to obtain the agent's pose $p_t$ (relative to pose at $t=0$). The agent then executes an action $a_t \sim \mathcal{A}$, where $\mathcal{A}$ consists of \code{move\_forward}, \code{turn\_left}, \code{turn\_right}, and \code{stop}. The agent is required to navigate within $d_s = 1.0\si{m}$ of the object and execute \code{stop} to successfully complete the task. The episode terminates when the agent executes \code{stop}, or if it exceeds a time budget of $T=500$ steps.

\subsection{Method Overview}

We propose Potential functions for ObjectGoal Navigation (PONI), a modular architecture for addressing ObjectNav (see~\cref{fig:approach}). Our model consists of three components. The semantic mapper uses RGB-D and pose sensor readings to build an allocentric semantic map ($m_t$) of the world capturing which objects are where on the ground plane (\cref{fig:approach}, left). The potential function network ($\pi_{pf}$) performs geometric and semantic reasoning on top of the semantic map and samples a long-term goal location $g_t$ to explore to find the goal object $o$. The local policy then navigates the agent towards the long-term goal using analytical path planning, and the process repeats until the episode ends. Our key contribution lies in the design and optimization of $\pi_{pf}$. Next, we discuss the individual components.

\subsection{Potential Function Network}
\label{sec:pfn}

The potential function network addresses the `where to look?' problem for an unseen goal object $o$. It uses the semantic map $m_t$ and the goal object $o$ to predict two potential functions that provide complementary information for ObjectNav (see~\cref{fig:approach}, center). The area potential function $U_t^a$ serves as a guide for efficient exploration, and helps find unexplored areas in the environment. Analogous to exploration rewards in RL~\cite{chaplot2020object,maksymets2021thda}, it provides a useful exploration bias for ObjectNav. When the semantic map is not informative (for example, at the start of the episode), $U_t^a$ is critical to quickly explore the environment and gather information. The object potential function $U_t^o$ serves as a guide to efficiently search for the goal object $o$. When the semantic map is sufficiently informative, $U_t^o$ is critical to perform semantic reasoning and quickly find the object. The potential function network combines these potential functions to sample the long-term goal, effectively trading-off between exploring the environment and finding the object. 

We first define the area and object potential functions which are analytically computed using a complete map of the environment during training, and then describe the potential function network architecture which learns to infer them given an incomplete map.

\begin{figure*}[t]
    \centering
    \includegraphics[width=1.0\textwidth,trim={0 57.5cm 33cm 0},clip]{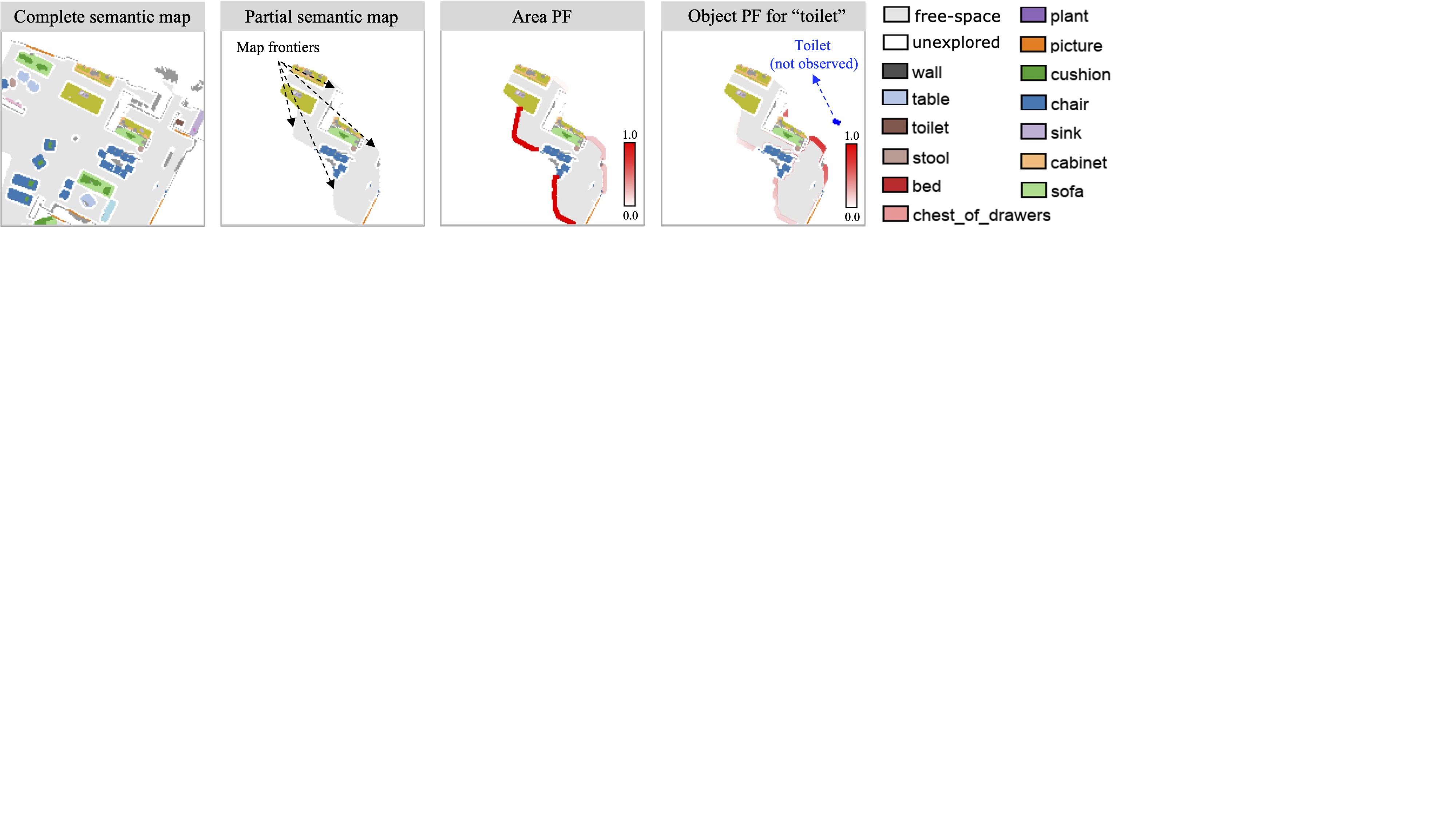}
    \vspace*{-0.25in}
    \caption{\small \textbf{An example of potential functions (PFs).} From left-to-right, we show the complete semantic map of the environment, the partial semantic map which only contains the parts observed by the agent, the area PF, and the object PF for the category `toilet'. The area and object PFs in red are defined on the frontiers and overlayed on the partial semantic map. The intensity of red indicates the value of the PF. While the area PF is high for frontiers that lead to more unexplored areas of the environment, the object PF is high for frontiers that are closest to the object. Note that during training these maps are augmented with random translation and rotation (see~\cref{sec:offline_learning}).}
    \label{fig:pf_examples}
\end{figure*}

In Fig.~\ref{fig:pf_examples}, we show an example semantic map and its corresponding potential functions. The `complete semantic map' is the full map of an environment, and the `partial semantic map' only contains the areas observed by an agent navigating in the environment. We define the area and object potentials only at the map frontiers, i.e., the edges between explored free-spaces (light-gray) and unexplored areas (white) on the partial semantic map (Fig.~\ref{fig:pf_examples}, column 2). Potential functions at the frontiers are sufficient for finding an unseen object since the path to any other unexplored location must pass through the frontier (by definition). In our experiments, we also find that predicting potential functions only at the frontiers is more effective than predicting it at all map locations (see~\cref{sec:results}).  \vspace{0.03in}

\texthead{Area potential function ($U_t^a$)} The area potential $U_t^a(f)$ at a frontier $f$ measures the amount of free-space left to explore beyond $f$, i.e., navigable cells which are unexplored in the partial semantic map. To calculate $U_t^a(f)$ on training data, we first group the unexplored free-space cells into connected components $C = \{c_1, \cdots, c_n\}$ using OpenCV~\cite{opencv_library}, and then associate each connected component $c_i$ to the map frontiers. A component $c$ is associated with frontier $f$ only if at least one pixel in $c$ is an 8-connected neighbor~\cite{wiki:Pixel_connectivity} of some pixel in $f$. For each frontier $f$, we can then calculate the area potential $U_t^a(f)$ as the sum of areas of connected components associated with $f$, normalized by the total free-space on the complete map.\footnote{For MP3D, we normalize by a fixed constant since the maps are huge.} To get the overall area potential function $U_t^a$ for the map, we set all non-frontier pixels to 0, and set a frontier's area potential to the corresponding frontier pixels (as shown in Fig.~\ref{fig:pf_examples}, column 3). \vspace{0.03in}

\texthead{Object potential function ($U_t^o$)} The object potential function $U_t^o$ for object $o_t$ is a function of the geodesic distance between a frontier location $x$ and $o_t$. \vspace{-0.05in}
\begin{equation}
    U_t^o(o_t, x) = \textrm{max}\bigg(1 - \frac{d_{g}(o_t, x)}{d_{\textrm{max}}},~0.0\bigg)
\end{equation}
where $d_g$ is the geodesic distance between $x$ and the $1.0\si{m}$ success zone surrounding the nearest object from category $o_t$ (similar to~\cite{batra2020objectnav}), and $d_{\textrm{max}}$ is the distance at which $U_t^o$ decays to $0$, selected via validation experiments. This object potential function facilitates efficient object search and is reminiscent of the heuristic in A* search~\cite{hart1968formal}. \vspace{0.03in} 

Next, we define the potential function network that infers the area and object potential functions from the partial semantic map. It consists of three components: the semantic map encoder $E$, the area potential decoder $D_a$, and the object potential decoder $D_o$, as shown in~\cref{fig:approach}.  \vspace{0.03in}

\texthead{Map encoder ($E$)} It extracts spatial features from the semantic map: $e_t = E(m_t)$ using a standard UNet encoder with 4 downsampling convolutional blocks~\cite{ronneberger2015u}. \vspace{0.03in} 

\texthead{Area potential decoder ($D_a$)} It predicts the area potential function conditioned on the encoder features: $U_t^a = D_a(e_t)$. We use a standard UNet decoder which consists of 4 upsampling convolutional blocks with a sigmoid activation function on the last layer. The output is a 1-channel map that represents the area potential at each map location. \vspace{0.03in}

\texthead{Object potential decoder ($D_o$)} It predicts the object potential function for all valid object categories conditioned on the encoder features: $U_t^o = D_o(e_t)$. We use a standard UNet decoder which consists of 4 upsampling convolutional blocks with a sigmoid activation function on the last layer. The output is a $N$-channel map representing the object potential for each object category ($1$ to $N$) at each location. To get the potential for a specific object category $o$, we select the corresponding map channel from $U_t^o$. \vspace{0.03in}

\texthead{Long-term goal sampling} We linearly combine the area and object potentials to obtain the overall potential which trades off between discovering unexplored areas and finding the object:  \vspace{-0.16in}
\begin{equation}
    \label{eqn:joint_potential}
    U_t = \alpha U_t^a + (1 - \alpha) U_t^o,
\end{equation}
where $\alpha=0.5$ is decided via validation experiments. To sample a long-term goal, we zero-out $U_t$ at all explored map locations (except frontiers) since we define the potentials only on the frontiers. Since the geometrically-calculated map frontiers can be noisy during navigation, we retain the predictions from the unexplored locations, providing the model some flexibility in deciding the frontier boundaries. We then sample the maximum location of the filtered $U_t$ as the long-term goal.

\subsection{Semantic Mapper}
The semantic mapper is responsible for aggregating the semantic information from individual RGB-D observations from time $0$ to $t$ into an allocentric semantic map $m_t$. We use the mapping procedure from a state-of-the-art semantic exploration method~\cite{chaplot2020object}. The depth observations are used to compute point-clouds that are registered to an allocentric coordinate system using the agent's poses ($p_0, \cdots, p_t$). Each point in the point-cloud is classified into one of $N$ object classes and $1$ background class by segmenting the corresponding RGB image using state-of-the-art segmentation models~\cite{he2017mask,jiang2018rednet}. The point-cloud is projected to the top-down map space by using differentiable geometric operations~\cite{henriques2018mapnet} to obtain the $(N + 2)\times M \times M$ semantic map $m_t$. Channels 1 and 2 correspond to obstacles and explored areas, and the rest contain the $N$ object categories.

\subsection{Analytical Local Policy} The local policy  $\pi_L$ navigates the agent to the long-term goal sampled by the potential function network. It uses the Fast Marching Method~\cite{sethian1999fast} to compute the shortest path from the current location $p_t$ to the long-term goal $g_t$ using the obstacle channel from the semantic map $m_t$. The local policy then takes deterministic actions to navigate the agent along this shortest path. This was found to be as effective as a learned local policy in prior work~\cite{chaplot2020object,chaplot2019learning}.

\subsection{Training for Potential Function Network}
\label{sec:offline_learning}

Our key insight is that the `where to look?' question for ObjectNav can be treated as a pure perception problem, and learned without any interactions in a simulated 3D environment. Specifically, we train the potential function network $\pi_{pf}$ on a dataset of semantic maps that are pre-computed from semantic annotations in 3D scene datasets~\cite{xia2018gibson,chang2017matterport3d}.\footnote{The same semantic annotations are used for most ObjectNav methods.} First, we project the semantic point-cloud annotations for a 3D scene to per-floor 2D semantic maps using the publicly available code from Semantic MapNet~\cite{cartillier2021semantic}. We then apply random translations and rotations to the map as a form of data augmentation. This gives an augmented and complete semantic map $m^c$ of the 3D scene with two channels for obstacles and explored regions, and $N$ channels for objects (see column 1 in Fig.~\ref{fig:pf_examples}).

From this complete semantic map $m^c$, we create a training data tuple ($m^p$, $U^a$, $U^o$) which consists of the partial semantic map, and the area and object potential functions (see columns 2 to 4 in Fig.~\ref{fig:pf_examples}). The partial map $m^p$ is a subset of $m^c$, and serves as a proxy for the semantic map an embodied agent would observe while navigating in the 3D environment. It is computed as follows. We initialize an all-zeroes `exploration mask' $m^e$ with the same size as $m^c$. We then randomly pick two free-space locations on the $m^c$ and compute the shortest path between them using a classical planner~\cite{sethian1999fast}. For each location $x$ on the shortest-path, we set a $S \times S$ square patch centered around $x$ to $1$ in $m^e$, indicating that these parts of the map have been explored. We copy values from the $m^c$ to $m^p$ only for locations that are set to `explored' in $m^e$. The remaining locations in $m^p$ are left unexplored. The distances to each object from the frontiers in $m^p$ are obtained using shortest-path planning~\cite{sethian1999fast,hart1968formal} on the complete map $m^c$. The area and object potential functions are computed as discussed in Sec.~\ref{sec:pfn}. 

Given a set of complete semantic maps $\{m^c_1, m^c_2, \cdots\}$ obtained from 3D scenes, we create the dataset $\mathcal{D}$ for training the potential function network as described above: \vspace{-0.02in}
\begin{equation}
    \mathcal{D} = \{(m^p_1, U^a_1, U^o_1), (m^p_2, U^a_2, U^o_2), \cdots\} \vspace{-0.02in}
\end{equation}
The potential function network is then trained to predict $(U^a, U^o)$ from $m^p$ using $\mathcal{D}$. Given a partial map $m^p$, the map encoder extracts features $e$, and the area and object potential decoders infer potentials $\hat{U}^a$ and $\hat{U}^o$, respectively. The models are trained end-to-end using the loss $L = L_a + L_c$, where $L_a$ and $L_c$ are pixel-wise mean-squared errors for the area and object potential functions. \vspace{-0.05in}
\begin{equation}
    L_a = \frac{1}{|\mathcal{F}|}\sum_{x \in \mathcal{F}}\big|\big|\hat{U}^a(x) - U^a(x)\big|\big|_2^2
\end{equation}
\begin{equation}
    L_c = \frac{1}{|\mathcal{F}|N}\sum_{x \in \mathcal{F}}\sum_{n = 1}^{N} \big|\big|\hat{U}^o(x, n) - U^o(x, n)\big|\big|_2^2
\end{equation}
Here, $\mathcal{F}$ is the set of frontier pixels, and $N$ is the number of object categories. The potential function network trained here is directly transferred to our PONI model in Fig.~\ref{fig:approach}.
\section{Experimental Setup}

We perform experiments on Gibson~\cite{xia2018gibson} and Matterport3D (MP3D)~\cite{chang2017matterport3d} datasets with the Habitat simulator~\cite{savva2019habitat}. Both Gibson and MP3D contain photorealistic 3D reconstructions of real-world environments. For Gibson, we use 25 train / 5 val scenes from the Gibson tiny split which have associated semantic annotations~\cite{armeni20193d}. For MP3D, we use the standard splits of 61 train / 11 val / 18 test scenes. 

We use the ObjectNav setup defined in~\cref{sec:objectnav_def}. These design choices are consistent with the CVPR 2021 ObjectNav Challenge~\cite{batra2020objectnav}. Note that the depth and odometer are noise-free in simulation. Only the semantic mapper relies on depth and pose, and this was shown to work well in the real world with noisy pose and depth in prior work~\cite{chaplot2020object}.

For Gibson experiments, we use the ObjectNav dataset from SemExp~\cite{chaplot2020object} which consists of 6 goal categories: `chair', `couch', `potted plant', `bed', `toilet', and `tv'. For MP3D experiments, we use the Habitat ObjectNav dataset~\cite{savva2019habitat} which consists of 21 goal categories (listed in supplementary). The semantic maps for training the potential function network use these categories as well.\vspace{0.05in}

\texthead{Evaluation metrics} We measure ObjectNav performance using four metrics. \textbf{Success} is the ratio of episodes where a method succeeded. \textbf{SPL} is the success weighted by the path length and measures the efficiency of the agent's path relative to the oracle shortest path length~\cite{batra2020objectnav}. \textbf{DTS} is the distance (in $\si{m}$) of the agent from the success threshold of the goal object at the end of the episode~\cite{batra2020objectnav}. \textbf{SoftSPL} is a softer version of SPL which measures efficiency based on the progress towards the goal (even with 0 success). It was introduced in the Habitat 2020 PointNav challenge.

\subsection{Baselines} \label{sec:baselines}

We use three types of baselines: non-interactive, end-to-end RL, and modular. \vspace{-0.1in}\\

\texthead{Non-interactive Baselines. } \\
\texthead{\code{BC}:} We use behavior cloning to train a ResNet-50 based recurrent policy~\cite{wijmans2019dd} which uses RGB-D, agent pose, and goal object category as inputs. \\ 
\texthead{\code{Predict-$\theta$}:} It classifies the direction to the nearest object. The directions from $0^\circ$ to $360^\circ$ are discretized into 8 classes. During ObjectNav, it navigates to the closest frontier along the predicted direction.\\
\texthead{\code{Predict-$xy$}:} It predicts the $(x, y)$ map location of the nearest object (same action space as~\cite{chaplot2020object}). During ObjectNav, it navigates to the predicted $(x, y)$ location. \\
\texthead{\code{Predict-$\mathcal{A}$}:} It classifies the navigation action to take to reach the nearest object along the shortest-path. The predicted action is executed at each step during ObjectNav.

The \code{Predict-*} baselines are obtained by changing the output parameterization of the potential function network in our PONI model (see~\cref{fig:approach}).  These are trained on the same dataset of semantic maps from~\cref{sec:offline_learning}.  \vspace{-0.1in}\\

\texthead{End-to-end RL Baselines. } \\
\texthead{\code{DD-PPO}~\cite{wijmans2019dd}:} It represents vanilla end-to-end RL with distributed training over several nodes. \\ 
\texthead{\code{Red-Rabbit}~\cite{ye2021auxiliary}:} It augments \code{DD-PPO} with multiple auxiliary tasks that improve sample efficiency and generalization to unseen environments. This was the winning entry to the Habitat ObjectNav challenge held at CVPR 2021.\\
\texthead{\code{THDA}~\cite{maksymets2021thda}:} It introduces `Treasure Hunt Data Augmentation' and improves the RL reward and model inputs, which results in better generalization to new scenes---at the time of submission, the state-of-the-art on MP3D. \vspace{-0.1in}\\

\texthead{Modular Baselines. } \\
\texthead{\code{SemExp}~\cite{chaplot2020object}:} This is the state-of-the-art modular method for ObjectNav. It uses RL-based interactive training to learn a policy for sampling long-term goals. It was the winning entry to the ObjectNav challenge held at CVPR 2020. \\
\texthead{\code{FBE}~\cite{yamauchi1997frontier}:} This is a classical frontier-based exploration approach that builds a 2D occupancy map of the world and navigates to the nearest map frontiers. When the semantic segmentation model detects the goal object, it navigates to the goal using an analytical local policy and executes stop. \\
\texthead{\code{ANS}~\cite{chaplot2019learning}:} This is a modular RL policy trained to maximize area coverage. It uses the same heuristic as \code{FBE} for goal detecting and stopping.

\code{FBE} and \code{ANS} perform goal-agnostic exploration and help benchmark the value of goal-driven behavior for ObjectNav.
For \code{DD-PPO}, \code{Red-Rabbit}, \code{THDA} and \code{SemExp}, we use the publicly available MP3D results on the Habitat ObjectNav challenge leaderboard. For \code{SemExp} on Gibson, and \code{ANS}, we evaluate the pre-trained models released by the authors.

\subsection{Implementation Details} 
On Gibson,  we finetune a COCO-pretrained Mask-RCNN~\cite{he2017mask} on images from the training split of Gibson tiny with $15$ object categories from~\cite{chaplot2020object}. On MP3D, we use a RedNet~\cite{jiang2018rednet} segmentation model that is trained in~\cite{maksymets2021thda}, and predict $21$ object categories. 

The potential function network uses a UNet-based encoder-decoder architecture~\cite{ronneberger2015u}. This model is trained on a dataset of semantic maps as described in~\cref{sec:offline_learning}. For Gibson, we extract 63 train / 13 val maps from each floor in Gibson tiny. For MP3D, we extract 153 train / 21 val maps from each floor. For each dataset, we pre-compute $400@000$ train / $1@000$ val $(m^p, U^a, U^o)$ tuples as discussed in \cref{sec:offline_learning}.  We train the model using PyTorch~\cite{pytorch} for 3 epochs. We use the Adam optimizer~\cite{kingma2014adam} with a learning rate of $0.001$ that is decayed by a factor of $10$ after 2 epochs. 

During ObjectNav transfer, we found it beneficial to sample a long-term goal with a frequency of T=1 steps for \code{PONI} (based on val performance). This is unlike prior modular methods~\cite{chaplot2019learning,chaplot2020object,ramakrishnan2020occupancy} which learn to sample long-term goals with T=25 steps. For \code{Predict-$\theta$} and \code{Predict-$xy$}, we found T=25 to work better. 
\section{Results}
\label{sec:results}


\begin{table}[t]

\begin{minipage}{\linewidth}
    \centering
    \resizebox{1.0\textwidth}{!}{
    \begin{tabular}{@{}l|ccc|cccc@{}}
    \toprule
                                              & \multicolumn{3}{c|}{Gibson (val)}     & \multicolumn{3}{c }{MP3D (val)}          \\ \cmidrule(l){2-4}\cmidrule(l){5-7} 
    Method                                    & Succ. \uarr &  SPL \uarr &  DTS \darr &  Succ. \uarr  & SPL \uarr &    DTS \darr \\ \midrule
    \code{BC}                                 &     12.2    &      8.3   &     3.90   &      3.8      &    2.1    &    7.5       \\
    \code{Predict-$\mathcal{A}$}              &     14.7    &     13.6   &     3.45   &      2.7      &    1.6    &    7.8       \\
    \code{Predict-$\theta$}                   &     69.9    &     35.7   &     1.44   &      29.0     &    10.6   &    5.7       \\
    \code{Predict-$xy$}                       &     66.9    &     34.2   &     1.69   &      29.4     &    10.7   &    5.5       \\ \midrule
    \code{DD-PPO}~\cite{wijmans2019dd}        &     15.0    &     10.7   &     3.24   &       8.0     &    1.8    &    6.9       \\
    \code{Red-Rabbit}~\cite{ye2021auxiliary}  &      -      &      -     &      -     & \tb{34.6}     &    7.9    &     -        \\
    \code{THDA}~\cite{maksymets2021thda}      &      -      &      -     &      -     &     28.4      &   11.0    &    5.6       \\ \midrule 
    \code{FBE}~\cite{yamauchi1997frontier}    &     64.3    &     28.3   &     1.78   &     22.7      &    7.2    &    6.7       \\
    \code{ANS}~\cite{chaplot2019learning}     &     67.1    &     34.9   &     1.66   &     27.3      &    9.2    &    5.8       \\
    \code{SemExp}~\cite{chaplot2020object}    &     71.7    &     39.6   &     1.39   &        -      &      -    &        -     \\ \midrule
    \code{PONI} (ours)                        & \tb{73.6}   & \tb{41.0}  & \tb{1.25}  &     31.8      & \tb{12.1} &  \tb{5.1}    \\ \bottomrule
    \end{tabular}
    }
    \vspace*{-0.1in}
    \caption{\textbf{ObjectNav validation results on Gibson and MP3D.} We train with 3 seeds and report the average performance. The missing results were not reported in the corresponding papers.}
    \label{tab:objectnav_results}
\end{minipage}
\vspace{0.05in}

\begin{minipage}{\linewidth}
    \centering
    \includegraphics[width=1.0\textwidth,clip,trim={0 3.75cm 5.5cm 0}]{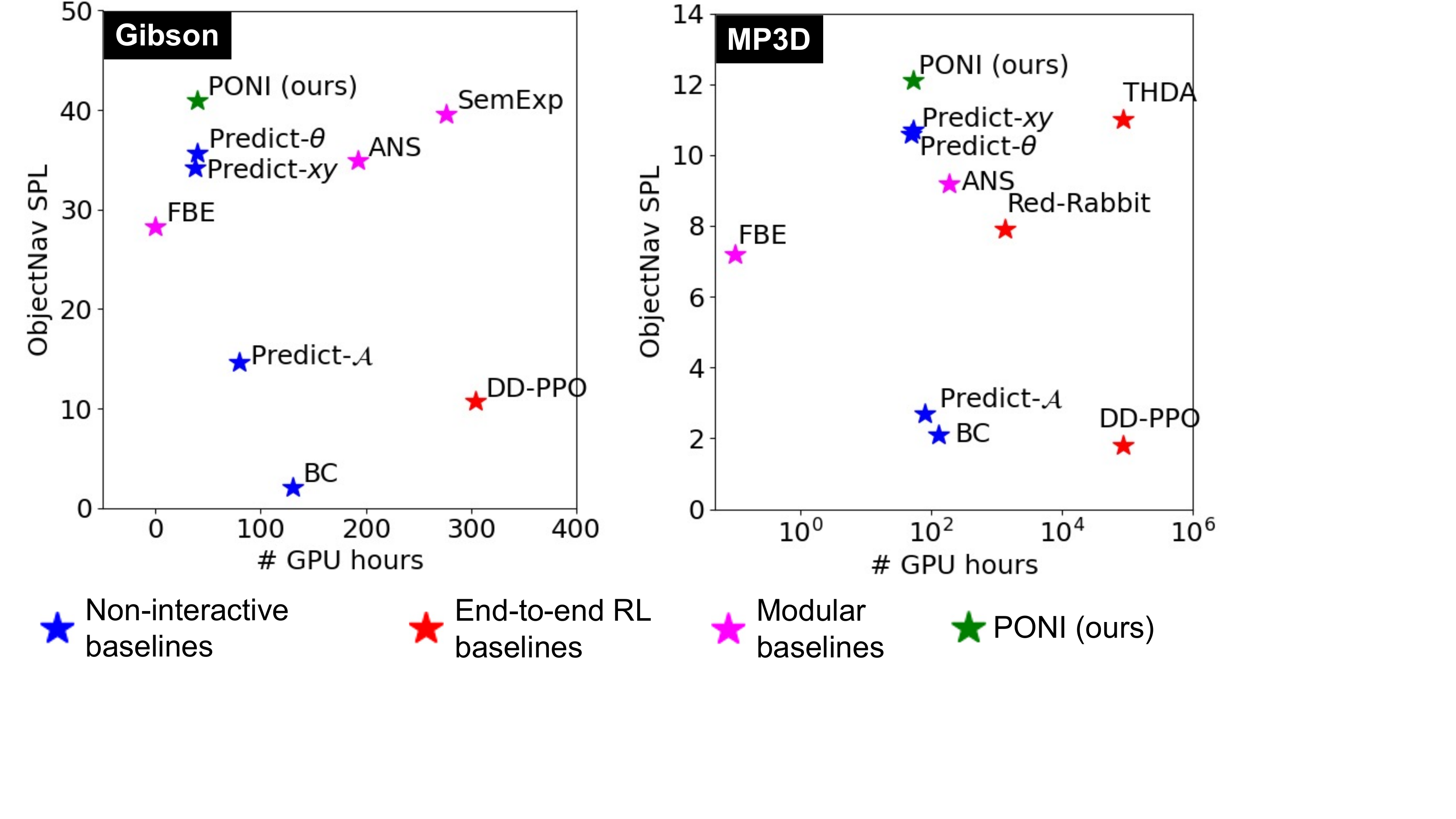}
    \vspace*{-0.2in}
    \captionof{figure}{\textbf{ObjectNav performance vs.~training cost.} We quantify training cost using $\#$ GPU hours used to train the model. PONI achieves state-of-the-art performance with up to $1@600\times$ lower training cost. Note: the MP3D plot uses a log-scale for X axis. }
    \label{fig:perf_vs_gpu_hours}
\end{minipage}
\end{table}

\begin{figure*}[t]
    \centering
    \includegraphics[width=1.0\textwidth,clip,trim={0 33.5cm 0 0}]{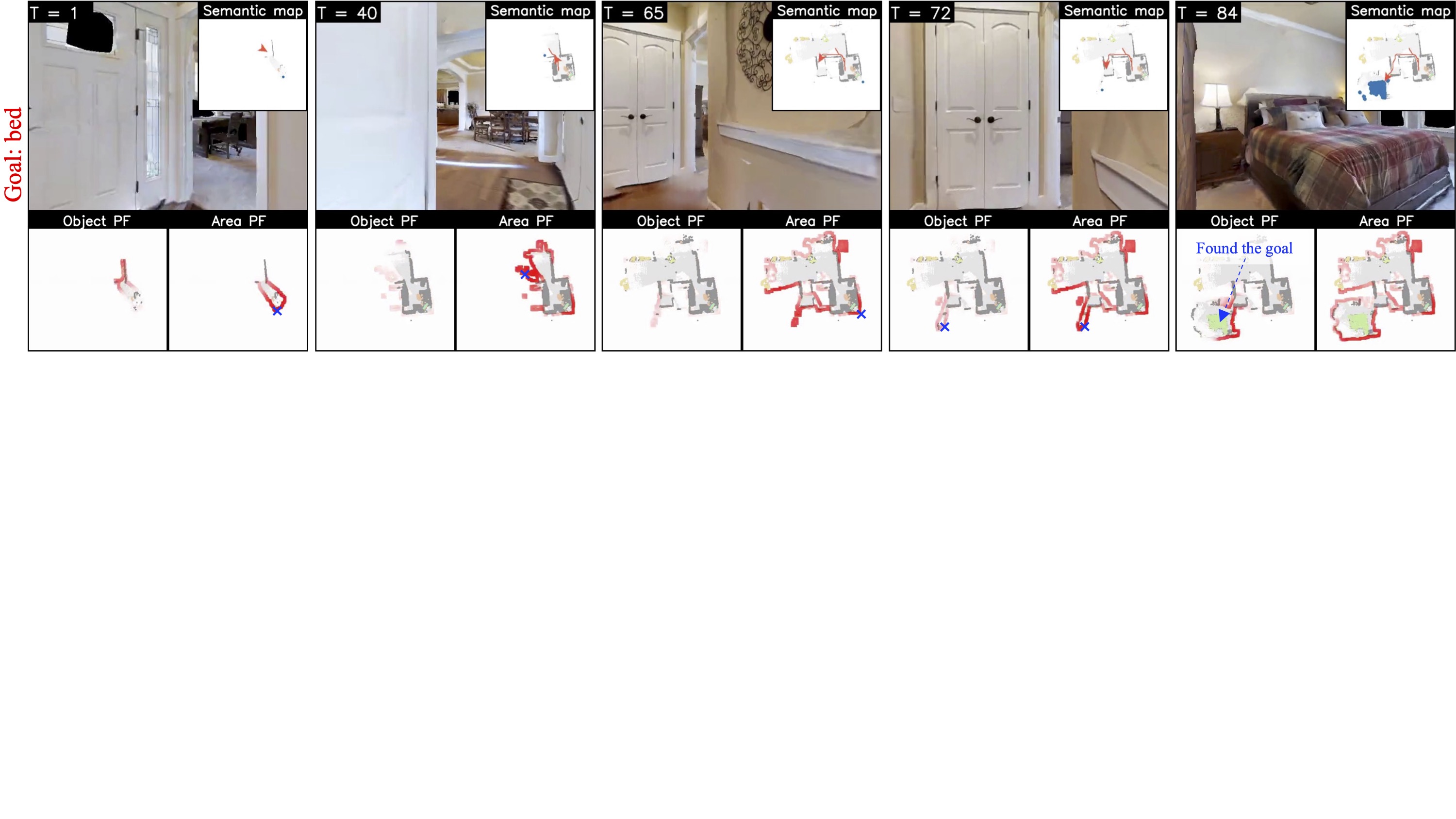}
    \vspace*{-0.30in}
    \caption{\textbf{Qualitative example of navigation using potential functions}. We visualize parts of an ObjectNav episode on Gibson (val), starting from T=1 until the agent finds the goal object (bed). For each step, we show the egocentric RGB view, the predicted semantic map, object and area potential functions. We indicate the maximum location that the agent navigates to using a blue cross on the PF map(s) responsible for the maximum. At the episode start (T=1 to 65), the agent is guided by the area PF which is high near frontiers leading to unexplored areas, allowing it to explore and gather information. The object PF plays a limited role here. After having gathered information, the model predicts higher object PF near the bedroom entrance at T=72, while the area PF remains high at multiple frontiers unrelated to the object location. The agent uses the new signal from the object PF to enter the bedroom and find the bed at T=84. This highlights the value of the two potential functions and how they are combined to perform ObjectNav. Please see supplementary for more examples.}
    \label{fig:pf_qual}
\end{figure*}

\cref{tab:objectnav_results} presents the ObjectNav performance on the Gibson and MP3D validation splits. We group the baselines into non-interactive (rows 1 - 4), end-to-end RL (rows 5 - 7), and modular methods (rows 8 - 10). \vspace{0.04in}

\textbf{Comparison to non-interactive baselines.} \code{PONI} is the best method for learning ObjectNav without interactions (see rows 1 - 4, 11 in~\cref{tab:objectnav_results}). On Gibson (val), \code{PONI} outperforms the next best non-interactive method (i.e., \code{Predict-$\theta$}) with $3.7\%$ higher success, $5.2\%$ higher SPL, and $0.19 \si{m}$ lower DTS. On MP3D (val), \code{PONI} outperforms the next-best method (i.e., \code{Predict-$xy$}) with $2.4\%$ higher success, $1.4\%$ higher SPL, and $0.4\si{m}$ lower DTS. Note that \code{PONI} is better than \code{Predict-$\theta$} and \code{Predict-$xy$} even though they are all trained on the same dataset of semantic maps with the same encoder backbone. This indicates that it is better to explicitly predict the area and object potential functions for navigation, instead of directly predicting where the objects are. \code{BC} and \code{Predict-$\mathcal{A}$} perform poorly suggesting that directly learning to classify shortest-path actions is inadequate. \vspace{0.04in}

\textbf{Comparison to end-to-end RL baselines.} \code{PONI} also outperforms the state-of-the-art in end-to-end RL (see rows 5 - 7 in~\cref{tab:objectnav_results}). \code{PONI} is significantly better than vanilla RL in \code{DD-PPO} on both Gibson (val) and MP3D (val). \code{Red-Rabbit} and \code{THDA} improve upon \code{DD-PPO} using techniques such as auxiliary tasks, data augmentation, and better reward-designs. When compared to these, \code{PONI} achieves more efficient navigation with $1-4\%$ higher SPL and competitive success rates on MP3D (val). \vspace{0.04in}

\textbf{Comparison to modular baselines.} \code{PONI} is the best modular method for ObjectNav (see rows 8 - 11 in~\cref{tab:objectnav_results}). \code{PONI} convincingly outperforms the goal-agnostic baselines \code{FBE} and \code{ANS} on all metrics and datasets, confirming the value of goal-oriented search for ObjectNav. \code{PONI} is also better than \code{SemExp}, the previous state-of-the-art modular method, on Gibson (val) even though they rely on the same semantic mapper and analytical local policy. This confirms our hypothesis that the `where to look?' question can be fundamentally treated as a perception problem and learned without any interactions. See~\cref{fig:pf_qual} for a qualitative visualization of \code{PONI}. \vspace{0.04in}

\textbf{Analysis of computational cost.} In~\cref{fig:perf_vs_gpu_hours}, we plot the ObjectNav SPL of different methods as a function of the computational cost for training. We quantify computational cost using the effective number of GPU hours (i.e., $\#$ GPUs $\times$ training time).\footnote{For end-to-end RL, we use the training cost reported in the papers.} See~\cref{fig:perf_vs_gpu_hours}. \code{PONI} achieves the state-of-the-art on Gibson (val) and MP3D (val) while having one of the lowest computational costs for training. In particular, \code{PONI} outperforms the prior SoTA on Gibson (\code{SemExp}) with $7\times$ lower training cost, and the prior SoTA on MP3D (\code{THDA}) with $1@600\times$ lower training cost. This highlights the value of treating `where to look?' as a perception problem.  \vspace{0.04in}

\textbf{Performance on Habitat Leaderboard.} We submitted our best-performing model to the Habitat challenge leaderboard. The results are in~\cref{tab:challenge_results}. At the time of submission, our method achieved the state-of-the-art relative to prior published entries, confirming our validation results. \vspace{0.04in}

\begin{table}[t]

\begin{minipage}{\linewidth}
    \centering
    \resizebox{1.0\textwidth}{!}{
    \begin{tabular}{@{}l|cccc@{}}
    \toprule
                                              & \multicolumn{4}{c}{MP3D (test-standard)}                  \\ \cmidrule(l){2-5}
    Method                                    &      SPL \uarr & SoftSPL \uarr &  DTS \darr &  Success \uarr \\ \midrule
    \code{PONI} (ours)                        &  \tb{8.82}     &  \tb{17.08}   &\tb{8.68}   &    20.01       \\
    \code{THDA}~\cite{maksymets2021thda}      &      8.75      &      16.96    &    9.20    &    21.08       \\
    \code{Red-Rabbit}~\cite{ye2021auxiliary}  &      6.22      &      12.14    &    9.14    &\tb{23.67}      \\
    \code{SemExp}~\cite{chaplot2020object}    &      7.07      &      14.50    &    8.82    &    17.85       \\
    \code{DD-PPO}~\cite{wijmans2019dd}        &      0.00      &       0.94    &   10.32    &     0.00       \\ 
    \bottomrule
    \end{tabular}
    }
    \vspace*{-0.1in}
    \caption{\textbf{Habitat ObjectNav challenge results}. We report the test-standard results from the top-performing \emph{published} methods on the EvalAI leaderboard (as of November 14th, 2021). \code{PONI} is the state-of-the-art on 3 out of 4 metrics. 
    }
    \label{tab:challenge_results}
\end{minipage}
\vspace{0.02in}

\begin{minipage}{\linewidth}
    \centering
    \resizebox{1.0\textwidth}{!}{
    \begin{tabular}{@{}cccc|ccc|ccc@{}}
    \toprule
    \multicolumn{4}{c|}{\textbf{\texttt{PONI}} ablations}&      \multicolumn{3}{c|}{Gibson (val)}              &            \multicolumn{3}{c}{MP3D (val)}              \\ \midrule
      $U^o$ &   F-only   &   $U^a$  &  GT$^\dagger$  &     Succ. \uarr      &     SPL \uarr   &     DTS \darr  &      Succ. \uarr     &     SPL \uarr   &     DTS \darr \\ \midrule
     \cm    &            &          &                &      58.8            &      34.9       &     2.18       &       30.5           &     11.6        &  \tb{5.1}     \\
     \cm    &    \cm     &          &                &      65.1            &      37.9       &     1.76       &       30.8           &     12.0        &      5.2      \\
            &    \cm     &   \cm    &                &      72.7            &      39.4       & \tb{1.20}      &       31.1           &     11.8        &      5.3      \\
     \cm    &    \cm     &   \cm    &                &  \tb{73.6}           &  \tb{41.0}      &     1.25       &   \tb{31.8}          & \tb{12.1}       &  \tb{5.1}   \\ \midrule
     \cm    &    \cm     &   \cm    &  \cm           &      86.5            &      51.5       &     0.76       &       58.2           &     27.5        &      3.4      \\ \bottomrule
    \end{tabular}
    }
    \vspace*{-0.1in}
    \caption{\textbf{Ablation study of \code{PONI}.} We study the impact of the object potential function $U^o$, area potential function $U^a$, the choice to define potential functions only at the frontiers (F-only), and ground-truth image segmentation (GT). $^\dagger$ This is privileged. \vspace*{-0.05in}}
    \label{tab:ablation_study}
\end{minipage}
\end{table}

\textbf{Ablation study.}  We perform an ablation study to understand the impact of different components of \code{PONI}. There are three key components that contribute to our performance: the object potential function ($U^o$), the area potential function ($U^a$), and the fact that they are defined only at the frontiers (F-only). We additionally study the impact of using ground-truth image segmentation (GT). In~\cref{tab:ablation_study} (rows 1-4), we compare the performance of our complete model with variants that have one or more components missing. The complete model with the 3 components achieves the best performance (row 4,~\cref{tab:ablation_study}). When $U^o$ is removed (row 3,~\cref{tab:ablation_study}), both success rate and SPL drop by a good margin, indicating the value of goal-oriented search within \code{PONI}. When $U^a$ is removed (row 2,~\cref{tab:ablation_study}), the performance drops even more, which shows the importance of exploration for ObjectNav, echoing findings from recent work that encourage exploration for ObjectNav via rewards~\cite{chaplot2020object,maksymets2021thda} and tethered policies~\cite{ye2021auxiliary}. In row 5, we augment our complete model with the ground-truth semantic segmentation (GT). We observe that the performance improves significantly on all cases. Since the image segmentation impacts semantic mapping and the stopping behavior for the local policy, segmentation failures are a major source of error for \code{PONI}. 

\section{Conclusion}

We presented PONI, a modular approach for ObjectNav. Our key idea is to treat `where to look for an unseen object?' purely as a perception problem and address it without any interactions. To this end, we proposed the potential function network, an encoder-decoder model that predicts two complementary potential functions to decide `where to look?' for a goal object. We proposed a novel strategy to train this model in a supervised fashion using a dataset of semantic maps obtained from 3D semantic annotations, unlike existing ObjectNav methods which design reward functions for RL-based policy learning. Through experiments on Gibson and Matterport3D, we demonstrated that our method achieves the state-of-the-art for ObjectNav while incurring significantly lower training cost. We hope that our work will spur future research into compute-efficient training for embodied navigation.
\section{Acknowledgements}

UT Austin is supported in part by the IFML NSF AI Institute, the FRL Cog.~Sci.~Consortium, and DARPA L2M. K.G is paid as a Research Scientist by Meta AI. We thank the CVPR reviewers and meta-reviewers for their valuable feedback and suggestions. We thank Vincent Cartiller for sharing image segmentation models for MP3D.

{\small
\bibliographystyle{ieee_fullname}
\bibliography{egbib}
}

\newpage
\clearpage

\setcounter{section}{0}
\setcounter{figure}{0}
\setcounter{table}{0}
\renewcommand{\thesection}{S\arabic{section}}
\renewcommand{\thetable}{S\arabic{table}}
\renewcommand{\thefigure}{S\arabic{figure}}

\section*{Supplementary Materials}

We now provide additional information about our experimental settings and supporting qualitative visualizations. Below is a summary of the sections in the supplementary:
\begin{itemize}
    \item (\S\ref{suppsec:limitations}) Limitations
    \item (\S\ref{suppsec:additional_experimental}) Additional experimental details
    \item (\S\ref{suppsec:non_interactive_baselines}) Non-interactive baseline implementation details
    \item (\S\ref{suppsec:masking_strategy}) Masking strategy for semantic maps
    \item (\S\ref{suppsec:action_costs}) Action costs for long-term goal sampling
    \item (\S\ref{suppsec:object_pf_influence}) Influence of object PF over time
    \item (\S\ref{suppsec:semantic_maps}) Examples of semantic maps
    \item (\S\ref{suppsec:potential_functions}) Examples of potential functions
    \item (\S\ref{suppsec:qualitative_trajectories}) Visualizing ObjectNav episodes
\end{itemize}
In our \href{https://vision.cs.utexas.edu/projects/poni/}{project website}, we visualize complete ObjectNav trajectories and provides an intuition of how \code{PONI} works. These are animated versions of the ObjectNav episodes visualized in Fig.~5 from the main paper, and~\cref{fig:pf_qual_supp} in the supplementary.

\section{Limitations}
\label{suppsec:limitations}

In Sec. 5 from the main paper, we discussed the benefits of our proposed PONI method both in terms of achieving state-of-the-art results on ObjectNav, as well as computational benefits during training. However, we would like to acknowledge some limitations of our approach. 

One of our main limitations is our reliance on the semantic map as the only source for deciding when an object is found (i.e., to execute STOP). As discussed in the ablation study from Sec. 5, our performance is sensitive to the image segmentation quality. The success rate goes down by $14.9\%$ in Gibson and $45.4\%$ on MP3D relative to the performance with ground-truth segmentation (see~\cref{tab:limitation_segm}). Note that our ratio of SPL to success remains relatively stable (only $6\%$ reduction on Gibson and $19\%$ on MP3D), indicating that our search efficiency is not affected significantly by segmentation errors. Unlike end-to-end RL methods which may learn to be robust to the sensory noise, we do not have an inbuilt mechanism to handle failures in segmentation. This limitation of interaction-free learning can potentially be addressed by using the latest advances in segmentation. Additionally, segmentation errors in simulation can be caused by reconstruction artifacts in the 3D scenes. Experimenting on higher quality scenes, or testing in the real world may address this limitation. 

We also rely on access to human-annotated semantic information in 3D scenes. While this is standard practice for most approaches in ObjectNav~\cite{chaplot2020object,maksymets2021thda,ye2021auxiliary,yang2018visual}, alternative strategies exist for learning ObjectNav without access to any ground-truth semantic annotations in 3D scenes~\cite{chang2020semantic}. Such self-supervised approaches have the potential to be more scalable than our supervised approach. However, to the best of our knowledge, there are no purely self-supervised methods that achieve state-of-the-art results for ObjectNav.

\begin{table}[t]

\begin{minipage}{\linewidth}
    \centering
    \resizebox{1.0\textwidth}{!}{
    \begin{tabular}{@{}l|ccc|cccc@{}}
    \toprule
                        & \multicolumn{3}{c|}{Gibson (val)}     & \multicolumn{3}{c }{MP3D (val)}          \\ \cmidrule(l){2-4}\cmidrule(l){5-7} 
    Method              & Succ.       &  SPL        &  SPL / Succ. &     Succ.     &    SPL      &  SPL / Succ. \\ \midrule
    \code{PONI} + GT-s  &   86.5      &   51.5      &   0.596      &     58.2      &   27.5      &    0.47      \\
    \code{PONI}         &   73.6      &   41.0      &   0.557      &     31.8      &   12.1      &    0.38      \\ \midrule
    Relative drop       &\red{-14.9\%}&\red{-20.4\%}&\red{-6.5\%}  & \red{-45.4\%} &\red{-56.0\%}&\red{-19.1\%}  \\
    \bottomrule
    \end{tabular}
    }
    \vspace*{-0.1in}
    \caption{Impact of segmentation errors on \code{PONI}'s ObjectNav performance. The first row shows performance with ground-truth segmentation. The last row shows the relative drop in the performance when we remove ground-truth segmentation.}
    \label{tab:limitation_segm}
\end{minipage}
\vspace{0.05in}
\end{table}
\section{Additional experimental details}
\label{suppsec:additional_experimental}

We provide additional information about the experiments to supplement the main paper. The Gibson ObjectNav dataset from~\cite{chaplot2020object} consists of $6$ object categories: `chair', `couch', `potted plant', `bed', `toilet', and `tv'. The train split episodes are generated on-the-fly during training from $25$ train scenes in Gibson tiny. The val split consists of $1@000$ episodes from $5$ val scenes in Gibson tiny. The MP3D ObjectNav dataset from the Habitat challenge consists of $21$ object categories: `chair', `table', `picture', `cabinet', `cushion', `sofa', `bed', `chest of drawers', `plant', `sink', `toilet', `stool', `towel', `tv monitor', `shower', `bathtub', `counter', `fireplace', `gym equipment', `seating', and `clothes'. The train / val splits consist of $263@2422$ / $2@195$ episodes from $61$ / $11$ MP3D scenes. We share these datasets publicly on our project website: {\footnotesize\url{https://vision.cs.utexas.edu/projects/poni/}}.
\section{Non-interactive baseline details}
\label{suppsec:non_interactive_baselines}

We provide more details about the non-interactive baselines from Sec.4.1 in the main paper.\vspace{0.2cm}\\
\texthead{\code{BC}:} We train a recurrent policy using behavior cloning. The policy consists of a ResNet-50 backbone for encoding RGB-D observations, and MLP layers to encode the agent's pose and goal object category. The outputs of these models are concatenated and fed to a 2-layer LSTM with 512-D hidden states to aggregate observations over time. The LSTM hidden states are used by a linear layer to predict a probability distribution over the set of agent actions. This is a standard policy architecture for recent navigation methods~\cite{wijmans2019dd,maksymets2021thda}. The idea in behavior cloning is to supervise the policy to classify the ground-truth action sampled by an expert at each step. We use the greedy shortest-path sampler from Habitat~\cite{savva2019habitat} to sample expert actions to the goal object. The model is trained using the cross-entropy loss. \vspace{0.2cm}\\ 
\texthead{\code{Predict-$\theta$}:} We modify the potential function network from Sec.~3.3 in the main paper to predict the direction to the nearest object from each category. We discretize the directions from $0^\circ$ to $360^\circ$ into 8 classes. The model uses the partial semantic map as input and predicts a $N \times 8$ array of direction probabilities for the $N$ object categories. The model is trained on the semantic maps dataset from Sec.~3.6 in the main paper with the cross-entropy loss per object category. During ObjectNav, we sample the most-likely direction to the goal object category, and navigate to the closest frontier along this direction.\vspace{0.2cm}\\
\texthead{\code{Predict-$xy$}:} We modify the potential function network from Sec.~3.3 to predict the $(x, y)$ map location of the nearest object from each category. The model uses the partial semantic map as input and regresses the normalized position values from $0$ to $1$ (same action space as~\cite{chaplot2020object}). The model is trained on the semantic maps dataset from Sec.~3.6 in the main paper with the mean-squared error loss per object category. During ObjectNav, we sample the predicted $(x, y)$ location as the long-term navigation goal.\vspace{0.2cm} \\
\texthead{\code{Predict-$\mathcal{A}$}:} We modify the potential function network from Sec.~3.3 to predict the low-level navigation action for reaching the nearest object from each category. The model uses the partial semantic map as input and classifies, per object category, the action for reaching the nearest object along the shortest-path. The model is trained on the semantic maps dataset from Sec.~3.6 in the main paper with the cross-entropy loss per object category. During ObjectNav, we sample the most-likely prediction action to reach the goal object.

\section{Masking strategy for semantic maps} \label{suppsec:masking_strategy} 

We described our strategy to sample exploration masks in Sec.~3.6 in the main paper, where we sampled random shortest paths and revealed a $3\si{m}\times 3\si{m}$ square patch around each location on the shortest-path. We now experiment with an alternative strategy where we reveal a viewing cone in-front of the agent, where we aim to mimic the agent's visibility in 3D space. The results are shown in \cref{supptab:masking_strategy}. We find that it performs comparably with the `square' strategy, which we use as the default option for all of our experiments.

\begin{table}[t]
\begin{minipage}{\linewidth}
    \centering
    \resizebox{0.75\textwidth}{!}{
    \begin{tabular}{@{}l|ccc@{}}
    \toprule
                           &   \multicolumn{3}{c }{MP3D (val)}          \\ \cmidrule(l){2-4}
    Method                 &  Succ. \uarr  &  SPL \uarr  &  DTS \darr   \\ \midrule
    \code{PONI} (square)   & 31.8     &\tb{12.1}    & \tb{5.1}     \\
    \code{PONI} (view-cone)& \tb{31.9}     &\tb{12.1}    & \tb{5.1}     \\
    \bottomrule
    \end{tabular}
    }
    \caption{We measure the impact of masking strategy for generating training samples during PF training. In `square', we unmask a $3\si{m} \times 3\si{m}$ square region centered around each shortest-path location on the semantic map. In `view-cone', we unmask a viewing cone in-front of the agent with $3\si{m}$ radius and $90^\circ$ field-of-view. Both strategies perform comparably on the MP3D (val) split. } 
    \label{supptab:masking_strategy}
\end{minipage}
\end{table}

\section{Action costs for long-term goal sampling}\label{suppsec:action_costs}

As described in Sec.~3.3 in the main paper, we sample long-term goals by selecting the maxima of the overall potential (see Eqn.~2). An alternative is to take into account the cost of navigating from the agent's location to each map location as well (i.e., an action cost). For example, when there are two locations with similarly valued potentials, the agent could choose to navigate to the nearer one. We incorporate this into PONI by adding a distance potential function $U^d$ that is $1.0$ at the agent's location and linearly decreases as we move away:
\begin{equation}
    U_t = \alpha U_t^a + \beta U_t^o + \gamma U_t^d,~~\text{where}~~\alpha + \beta + \gamma = 1.
\end{equation}

The constants $\alpha, \beta, \gamma$ are determined through a grid-search over MP3D (val). We compare the best results from this grid-search (\code{+ act-cost}) with our current method in \cref{supptab:action_costs}. PONI does not benefit from adding the action costs. Based on our qualitative analysis, we find that the PONI agent typically continues to explore a single frontier sufficiently before moving away to other frontiers. Therefore, prioritizing the best frontier at all times (regardless of how far away it is) works well in practice.

\begin{table}[t]
\begin{minipage}{\linewidth}
    \centering
    \resizebox{0.8\textwidth}{!}{
    \begin{tabular}{@{}l|ccc@{}}
    \toprule
                           &   \multicolumn{3}{c }{MP3D (val)}          \\ \cmidrule(l){2-4}
    Method                 &  Succ. \uarr  &  SPL \uarr  &  DTS \darr   \\ \midrule
    \code{PONI}            & \tb{31.8}     &\tb{12.1}    & \tb{5.1}     \\
    \code{PONI + act-cost} &     30.3      &    11.6     &     5.3      \\
    \bottomrule
    \end{tabular}
    }
    \caption{We measure the impact of using action costs to sample long-term goals for PONI.}
    \label{supptab:action_costs}
\end{minipage}
\end{table}

\section{Influence of object PF over time} \label{suppsec:object_pf_influence} 

\begin{figure}[t]
    \centering
    \includegraphics[width=0.5\textwidth,clip,trim={0 5.2cm 8cm 0}]{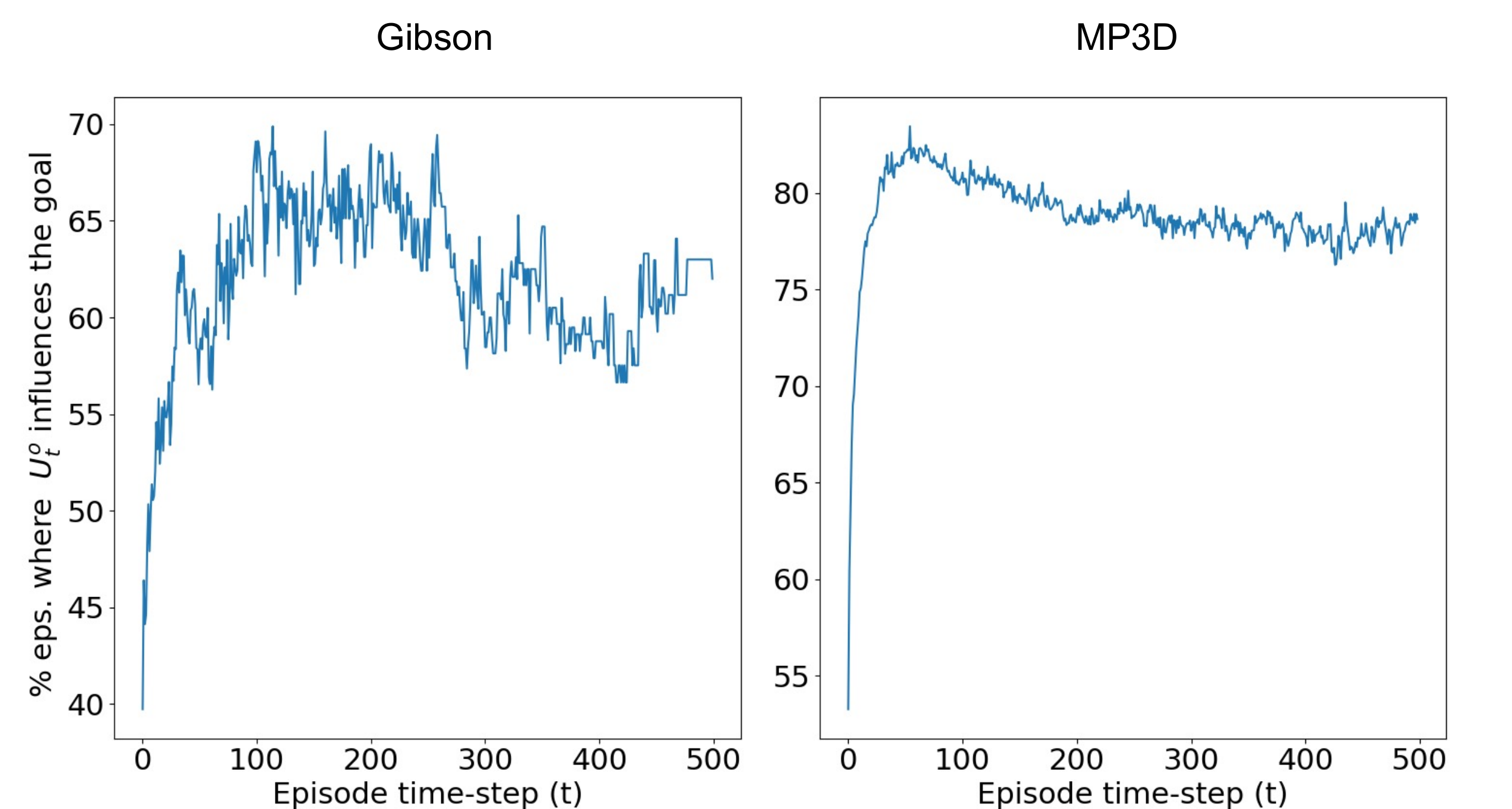}
    \caption{\textbf{Influence of object PF on action selection}: The plots show the percentage of of episodes where the selected goal location was influenced by the object PF (y-axis) at a given time-step of an episode (x-axis). The contribution of the object PF is higher during later stages of the episode.}
    \label{suppfig:object_influence}
\end{figure}

In Fig.~5 from main and~\cref{fig:pf_qual_supp}, we qualitatively demonstrated that the agent explores using the area PF in early stages of the episode, and then uses the object PF to find objects. We now quantitatively demonstrate this. In \cref{suppfig:object_influence}, we plot the influence of the object PF on the goal location selection at a given time $t$ on Gibson (val) and MP3D (val), i.e, the percentage of episodes where the selected goal location differs from the maxima of the area PF at $t$ (by atleast $1\si{m}$ euclidean distance). The contribution of the object PF is higher in the later stages of the episode, after sufficient information has been gathered. This is intuitive: we cannot anticipate unseen objects without sufficient context on the map. 

\section{Examples of semantic maps}
\label{suppsec:semantic_maps}
\begin{figure*}
    \centering
    \includegraphics[width=0.95\textwidth,trim={0 3cm 0 0},clip]{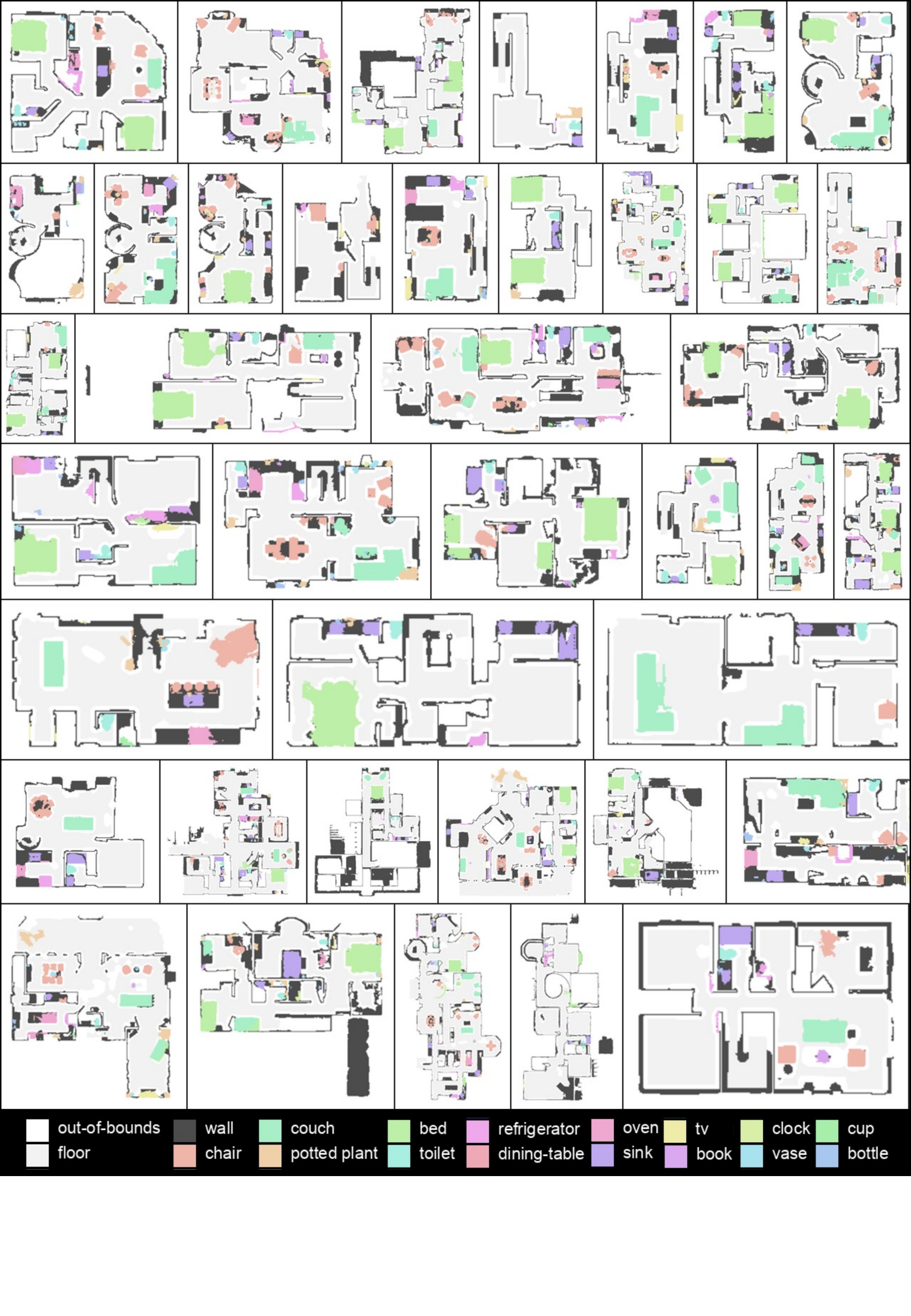}
    \caption{Examples of semantic maps from Gibson. The maps contain objects from up to 15 object categories (legend on the last row). }
    \label{fig:gibson_semantic_maps}
\end{figure*}

\begin{figure*}
    \centering
    \includegraphics[width=0.95\textwidth,trim={0 2.3cm 0.0cm 0},clip]{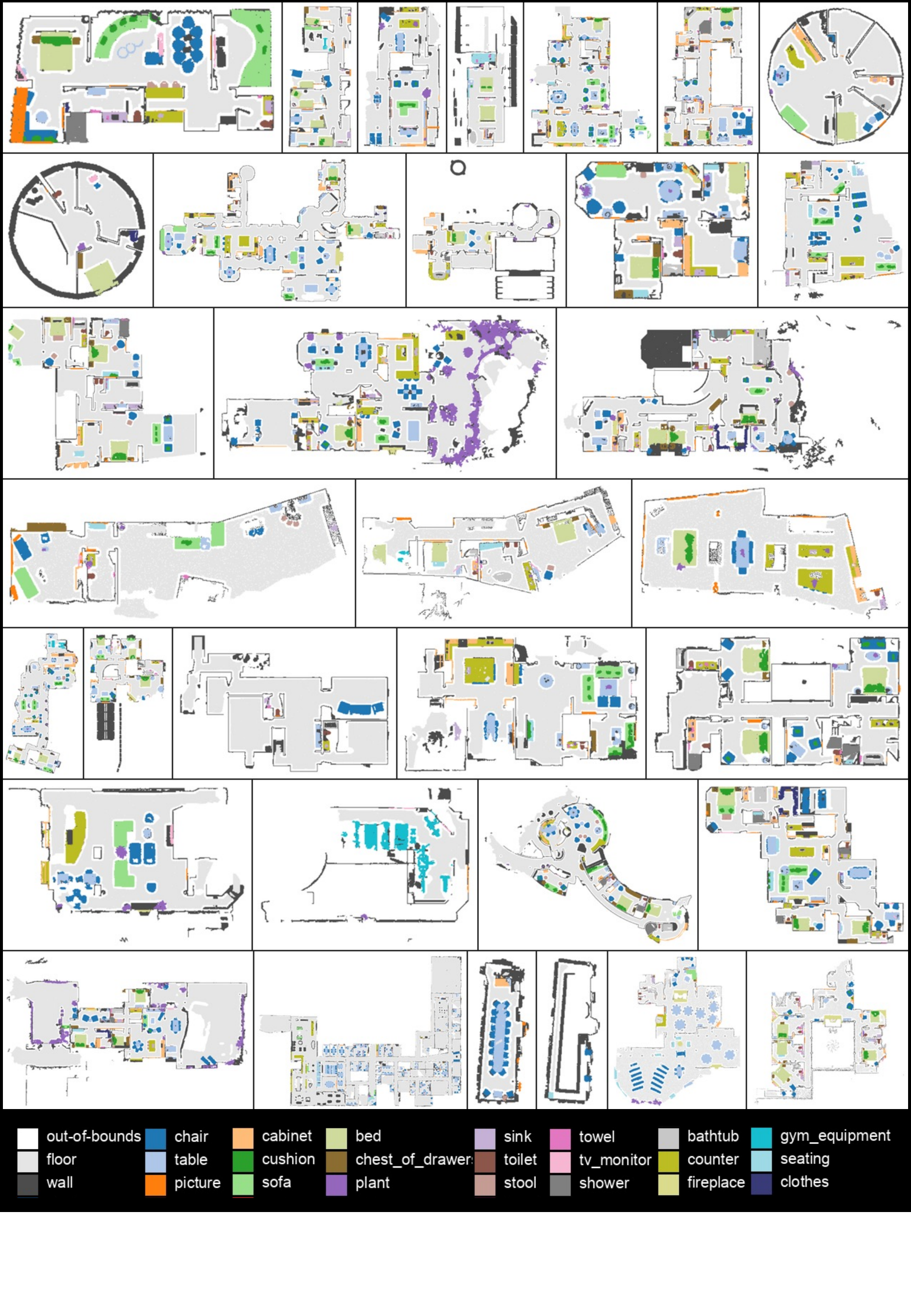}
    \caption{Examples of semantic maps from MP3D. The maps contain objects from up to 21 object categories (legend on the last row). }
    \label{fig:mp3d_semantic_maps}
\end{figure*}

We show examples from the semantic map datasets we used for training the potential function network in~\cref{fig:gibson_semantic_maps,fig:mp3d_semantic_maps}. The Gibson semantic maps contain up to 15 object categories of which 6 categories are goal categories (same as~\cite{chaplot2020object}). The MP3D semantic maps contain up to 21 object categories of which all are goal categories (same as the Habitat challenge~\cite{batra2020objectnav}). These maps are computed by performing an orthographic projection of the 3D point-cloud annotations (following~\cite{cartillier2021semantic}). In addition to the pipeline from~\cite{cartillier2021semantic}, we perform additional pre-processing to obtain per-floor maps. Specifically, we segment the 3D semantic point-cloud from Gibson~\cite{armeni20193d} and MP3D~\cite{chang2017matterport3d} annotations into different floors. We do this by loading each scene into Habitat~\cite{savva2019habitat}, identifying the navigable points and clustering them along the Y-coordinate to automatically discover the number of floors and their extents using DBScan~\cite{Ester1996ADA}. We then perform orthographic projection independently for each floor of the scene.

\section{Examples of potential functions}
\label{suppsec:potential_functions}

\begin{figure*}
    \centering
    \includegraphics[width=\textwidth,trim={0 27.5cm 6.0cm 0},clip]{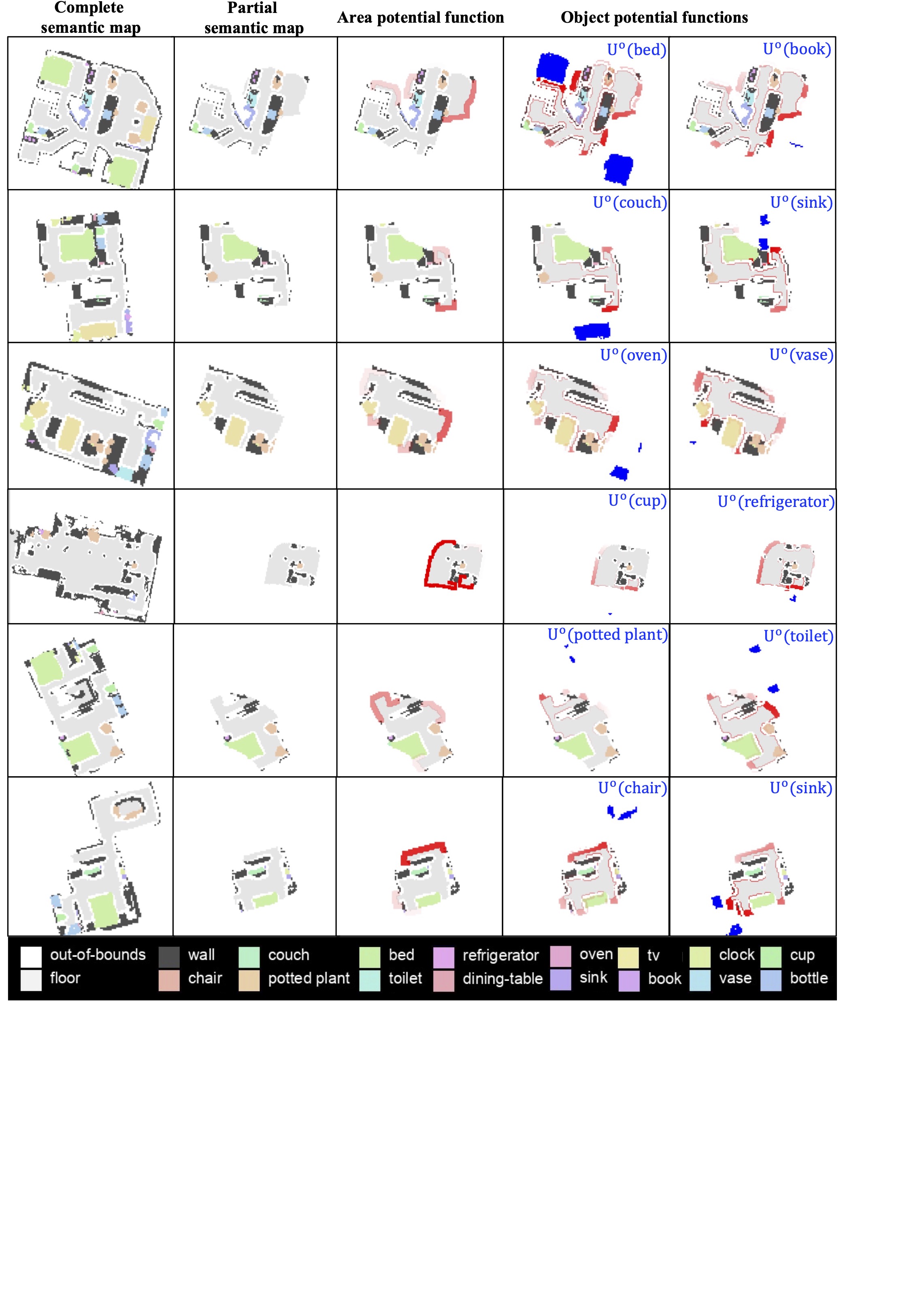}
    \caption{\textbf{Examples of potential functions from Gibson.} On each row, we show the complete semantic map, partial semantic map, the area potential function, and object potential functions for two unseen objects (from left to right). The potential functions are computed at the map frontiers using the analytical procedure described in Sec. 3.3 from the main paper. Both the potential functions range from 0.0 to 1.0, which the intensity of red indicating the strength of the potential function (1.0 is highest intensity). For the object potential function, we state the object category on the top-right corner of the map, and also highlight the spatial locations on the map in bright blue.}
    \label{fig:gibson_potential_fns}
\end{figure*}

\begin{figure*}
    \centering
    \includegraphics[width=\textwidth,trim={0 30cm 6cm 0},clip]{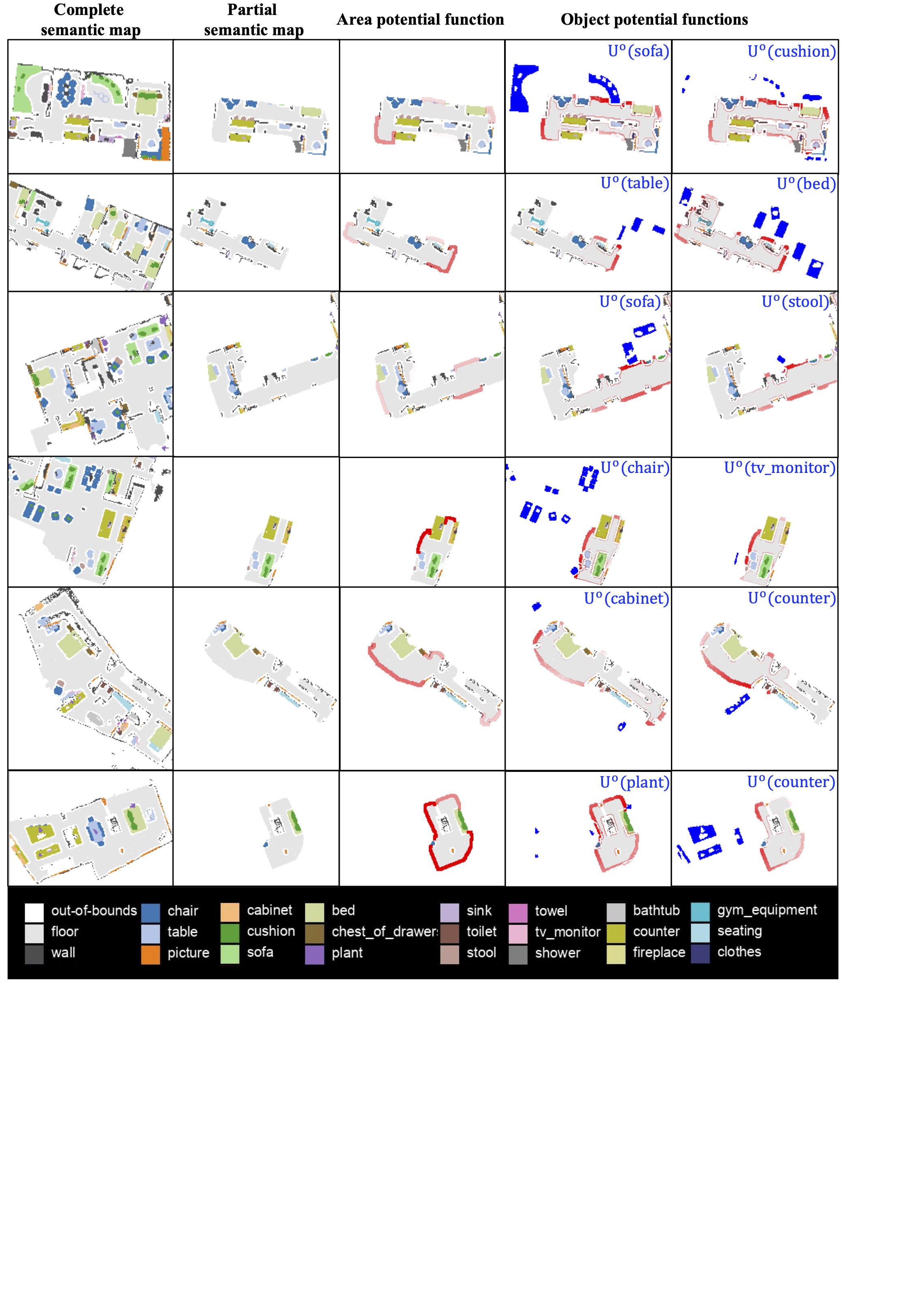}
    \caption{\textbf{Examples of potential functions from MP3D.} On each row, we show the complete semantic map, partial semantic map, the area potential function, and object potential functions for two unseen objects (from left to right). The potential functions are computed at the map frontiers using the analytical procedure described in Sec. 3.3 from the main paper. Both the potential functions range from 0.0 to 1.0, which the intensity of red indicating the strength of the potential function (1.0 is highest intensity). For the object potential function, we state the object category on the top-right corner of the map, and also highlight the spatial locations on the map in bright blue.}
    \label{fig:mp3d_potential_fns}
\end{figure*}

In Fig. 3 from the main paper, we showed an example of potential functions from MP3D. We now show more examples of such potential functions for Gibson and MP3D in~\cref{fig:gibson_potential_fns,fig:mp3d_potential_fns}.

\section{Visualizing ObjectNav episodes}
\label{suppsec:qualitative_trajectories}

\begin{figure*}[t]
    \centering
    \includegraphics[width=1.0\textwidth,clip,trim={0 7.5cm 0 0}]{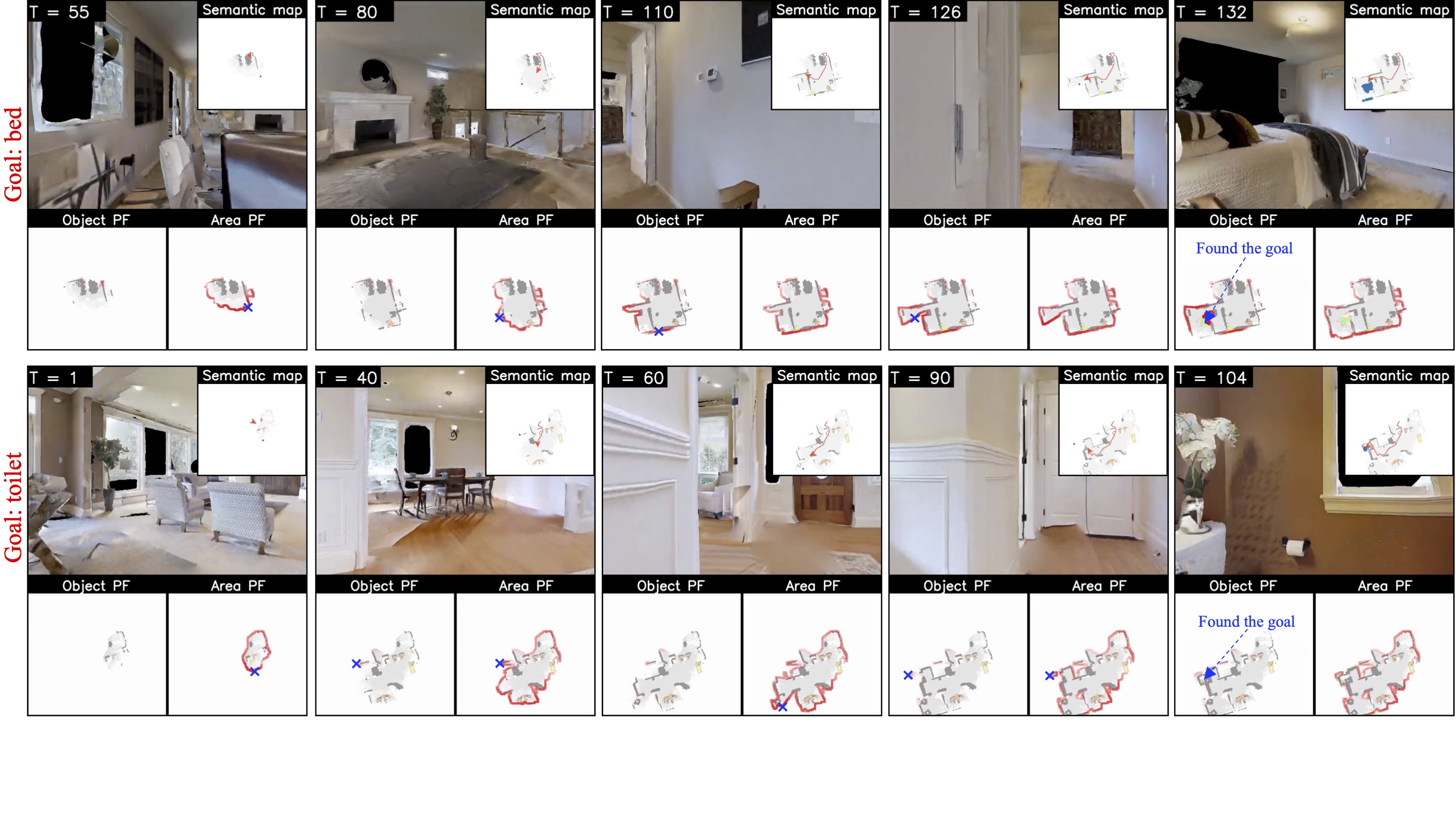}
    \caption{\textbf{Qualitative examples of navigation using potential functions}. On each row, we visualize parts of an ObjectNav episode on Gibson (val) as the agent searches for and finds the goal object. For each step, we show the egocentric RGB view, the predicted semantic map, object and area potential functions (PFs). We indicate the maximum location that the agent navigates to using a blue cross on the PF map(s) responsible for the maximum. \textbf{Row 1:} The agent searches for a bed in an unexplored scene. In the first several steps of the episode (T=1 until 110), the agent is guided by the area PF which is high near frontiers leading to unexplored areas, allowing it to explore efficiently and gather information. The object PF plays a limited role here. After having explored sufficient parts of the environment, the model predicts high object PF at two frontiers (one of which corresponds to the bedroom entrance), while the area PF remains high at multiple frontiers unrelated to the object location. Guided by the signal from the object PF, the agent starts entering the bedroom at T=126, and eventually finds the goal at T=132. \textbf{Row 2:} The agent searches for a toilet in an unexplored scene. In the initial steps of the episode (T=1 to T=60), the agent is primarily guided by the area PF to explore the scene and gather information. At T=90, the object PF activates near the toilet room entrance. Note that while the absolute value of the object PF is not very high, it is sufficient to bias the overall PF towards the goal (and away from other frontiers). This is critical since the area PF has high values along multiple frontiers, while the object PF focuses on frontiers that could lead to the object. The agent follows this signal to eventually reach the goal at T=104. These examples highlight the value of the two potential functions and how they are combined to perform ObjectNav.}
    \label{fig:pf_qual_supp}
\end{figure*}

In Figure 5 from the main paper, we qualitatively visualized an episode showing how the potential functions are used to perform ObjectNav. We provide two additional examples in~\cref{fig:pf_qual_supp}.

\end{document}